\documentclass[]{dreamvu_preamble/dreamvu}
\pdfoutput=1

\usepackage{verbatim}
\usepackage{amsmath,amsfonts,amssymb}
\usepackage{xspace}
\usepackage{url}
\usepackage{array}
\usepackage{tabularx}
\usepackage{adjustbox}
\usepackage{wrapfig}
\usepackage{lmodern}
\usepackage{enumitem}
\usepackage{cleveref}
\usepackage{comment}
\usepackage{float}
\usepackage{tikz}
\usetikzlibrary{arrows.meta,shapes,positioning,calc}
\usepackage{pgfplots}
\pgfplotsset{compat=1.18}
\usepackage{xcolor}
\definecolor{darkgreen}{rgb}{0.0, 0.5, 0.0}

\newcommand{\nbf}[1]{\noindent\textbf{#1.}~}

\newcommand{\paperTitle}{RetailSMV: Exocentric vs.\ Egocentric Adaptation of Foundation Video World Models in Retail}

\newtoggle{arxiv}
\toggletrue{arxiv}

\title{RetailSMV: Exocentric vs.\ Egocentric Adaptation of Foundation Video World Models in Retail}
\renewcommand{\titlelist}{{\LARGE\sffamily\bfseries \paperTitle}}

\author{DreamVu\protect\footnotemark}

\abstract{Foundation video diffusion models are increasingly viewed as world simulators for embodied agents, yet their pretraining on internet-scale generic video leaves them poorly aligned with real-world deployment domains. We study parameter-efficient adaptation of a pretrained foundation video world model to retail scenes: when synchronized egocentric and exocentric video of the same activity are available, which viewpoint of training data produces the strongest adapted model?

We introduce \textbf{RetailSMV} (\emph{Retail Synchronized Multi-View}), a corpus of $32{,}105$ captioned retail clips from five supermarkets with synchronized ego/exo capture from the \emph{store-staff} perspective (stocking, arranging, weighing, managing supply carts, scanning at checkout), rather than the customer-centric framing of prior retail video corpora, and train three matched Low-Rank Adaptation (LoRA) configurations of Cosmos3-Nano (egocentric-only, exocentric-only, combined) under identical hyperparameters. On a $200$-clip held-out test set evaluated with seven complementary metrics under a strict paired statistical protocol, exocentric-only adaptation matches or exceeds combined adaptation on six of seven point estimates and is significantly better on LPIPS, PSNR, and DreamSim, despite training on only $15{,}985$ exocentric clips (versus $32{,}105$ for combined). A symmetric paired comparison further shows that adding exocentric data to egocentric-only training \emph{helps} while adding egocentric data to exocentric-only training \emph{hurts}. The absolute adaptation gap is largest at the shortest rollout time, identifying the near-horizon prediction window as the regime in which adaptation is most beneficial.
}

\date{\today}

\begin{document}

\maketitle

\footnotetext{A detailed list of contributors and acknowledgments can be found in \cref{sec:appendix_contributors} of this paper.}

\begin{figure}[H]
\centering
\includegraphics[width=\linewidth]{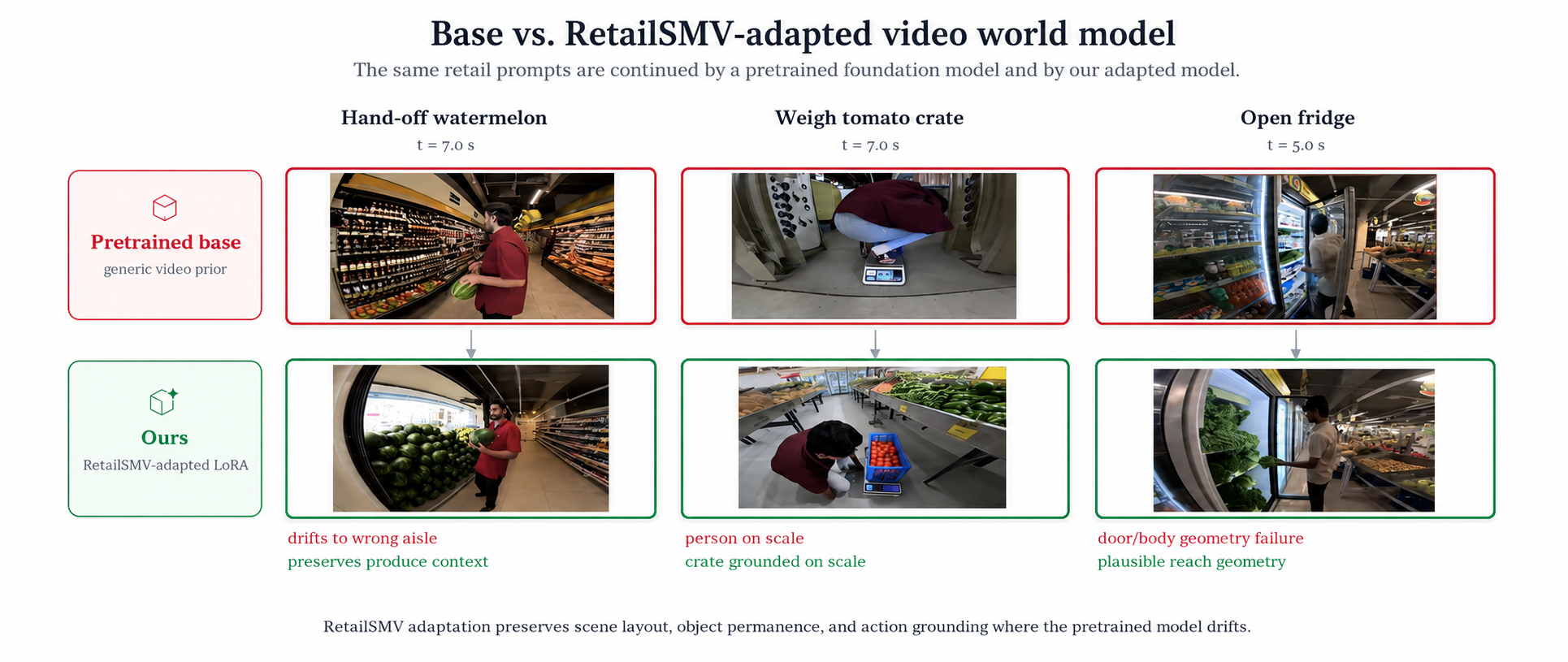}
\caption{\textbf{Base vs.\ RetailSMV-adapted video world model.} The same retail prompts are continued by the pretrained Cosmos3-Nano foundation model (top, red) and by our RetailSMV-adapted LoRA (bottom, green) under identical inference settings. RetailSMV adaptation preserves scene layout (hand-off watermelon), action grounding (weigh tomato crate), and physical geometry (open fridge) where the pretrained baseline drifts.}
\label{fig:teaser}
\end{figure}

\section{Introduction}
\label{sec:intro}

A video world model takes an observation history (typically a text description, an image, or a short video clip) and predicts a plausible video continuation. Modern video diffusion systems such as Sora~\cite{sora2024}, Stable Video Diffusion~\cite{svd2023}, Movie Gen~\cite{moviegen2024}, and NVIDIA Cosmos3-Nano and Cosmos-Predict~2.5~\cite{cosmos_paper} now produce coherent multi-second video from text prompts and have been framed as world simulators for embodied agents~\cite{ha2018worldmodels,gao2025worldmodel}. The promise is that a faithful world model could let an embodied agent plan, evaluate counterfactuals, generate training data, and verify policies before acting in the physical world. We discuss the broader landscape of video diffusion, world models, and embodied deployment in \cref{sec:related}.

That promise depends on domain alignment. Foundation video models are pretrained on internet-scale generic video (entertainment content, vlogs, dashcam footage, and robot demonstrations) which may underrepresent the structured visual vocabulary of specific deployment domains. Retail environments are a clear example: dense product shelving, narrow parallel aisles, repetitive geometry at multiple scales, distinctive signage and end-caps, and multi-person dynamics around carts and checkout counters. A retail world model must render scenes built from this vocabulary and continue shopping behaviors consistent with the physical and behavioral regularities of supermarket environments. Without explicit adaptation, generated retail scenes drift toward generic interiors, and predicted human motion drifts away from realistic shopping within a few seconds of rollout.

This paper studies parameter-efficient adaptation of a pretrained foundation video world model to retail scenes via Low-Rank Adaptation (LoRA)~\cite{lora}, and introduces the dataset that makes such a study possible. We use Cosmos3-Nano~\cite{cosmos_paper} as the base model, a 16-billion-parameter text-and-image-conditioned video diffusion transformer that is, to our knowledge, one of the largest openly released foundation video models with a documented post-training recipe. As training data we introduce \textbf{RetailSMV} (\emph{Retail Synchronized Multi-View}), a retail video corpus of $32{,}105$ captioned clips collected across five real-world supermarkets. RetailSMV provides \emph{synchronized} ego/exo capture of \emph{store-staff} operational work---stocking, arranging, weighing and labelling produce, carrying crates, pushing supply carts, scanning items at the checkout, and assisting customers---rather than the customer-side behaviour that prior retail video corpora capture. Every work episode is recorded simultaneously by a head-mounted egocentric camera worn by the staff member and a fixed exocentric scene camera that observes the same activity from a third-person perspective. To our knowledge no prior dataset combines a real-world retail deployment domain with synchronized ego/exo capture of the same activity at the scale required to fine-tune a foundation video diffusion model; this combination is what enables the controlled view-stratified study reported here.

\begin{figure*}[!t]
    \centering
    \resizebox{\textwidth}{!}{%
    \begin{tikzpicture}[
        sensorbox/.style={draw=blue!50!black, rounded corners=4pt,
            minimum height=1.0cm, minimum width=2.8cm, align=center,
            font=\small, line width=0.7pt},
        corpusbox/.style={draw=black!40, rounded corners=4pt,
            minimum height=1.0cm, minimum width=2.8cm, align=center,
            font=\small, line width=0.7pt},
        configbox/.style={draw, rounded corners=4pt,
            minimum height=1.3cm, minimum width=3.2cm, align=center,
            font=\small\bfseries, line width=1pt},
        modelbox/.style={draw=black!55, rounded corners=5pt,
            minimum height=2.8cm, minimum width=3.3cm, align=center,
            font=\small, line width=1pt, fill=yellow!8},
        evalpanel/.style={draw=black!55, rounded corners=6pt,
            inner sep=0.30cm, line width=1pt, fill=red!4,
            align=left, font=\scriptsize, text width=4.5cm},
        arrow/.style={-{Stealth[length=2.2mm,width=1.6mm]},
            line width=0.7pt, draw=black!75},
        arrowstrong/.style={-{Stealth[length=2.8mm,width=2.0mm]},
            line width=1.1pt, draw=black!85},
        colheader/.style={font=\small\itshape, text=black!60, anchor=south},
        colfooter/.style={font=\scriptsize\itshape, text=black!55,
            anchor=north},
    ]

    \def\cA{0}    \def\cB{4.5}  \def\cC{9.5}  \def\cD{14.4} \def\cE{19.6}

    \node[colheader] at (\cA, 2.7) {Synchronized capture};
    \node[colheader] at (\cB, 2.7) {Multi-view corpus};
    \node[colheader] at (\cC, 2.7) {View-stratified LoRA configurations};
    \node[colheader] at (\cD, 2.7) {Foundation model};
    \node[colheader] at (\cE, 2.7) {Held-out evaluation};

    \node[sensorbox, fill=blue!10]   at (\cA,  1.0) (ego) {Egocentric \\
        \scriptsize head-mounted camera};
    \node[sensorbox, fill=teal!12]   at (\cA, -1.0) (exo) {Exocentric \\
        \scriptsize fixed scene camera};
    \node[colfooter] at (\cA, -2.4) {5 supermarkets};

    \node[corpusbox, fill=blue!8]  at (\cB,  1.0) (egocorp)
        {Ego clips \\ $n{=}16{,}120$};
    \node[corpusbox, fill=teal!10] at (\cB, -1.0) (exocorp)
        {Exo clips \\ $n{=}15{,}985$};
    \node[colfooter] at (\cB, -2.4) {$32{,}105$ captioned clips};

    \node[configbox, draw=blue!50!black,   fill=blue!15]
        at (\cC,  1.8) (cfg1) {Egocentric-only \\
            \footnotesize\mdseries $n{=}16{,}120$};
    \node[configbox, draw=purple!55!black, fill=purple!12]
        at (\cC,  0.0) (cfg3) {Combined \\
            \footnotesize\mdseries $n{=}32{,}105$};
    \node[configbox, draw=teal!50!black,   fill=teal!15]
        at (\cC, -1.8) (cfg2) {Exocentric-only \\
            \footnotesize\mdseries $n{=}15{,}985$};
    \node[colfooter] at (\cC, -2.9) {identical hyperparameters,
        identical optimization budget};

    \node[modelbox] at (\cD, 0.0) (model) {%
        \textbf{Cosmos3-Nano} \\[1pt]
        \scriptsize 16B-param DiT \\[2pt]
        \scriptsize LoRA rank $32$, $\alpha{=}64$\\
        \scriptsize $3{,}000$ steps \\
        \scriptsize $480{\times}832$, $81$ frames};

    \node[evalpanel] at (\cE, 0.0) (evalp) {%
        \textbf{Test set}\;200 stratified clips (100 ego, 100 exo)\\[5pt]
        \textbf{Metric suite}\\
        \textit{Per-pixel:} LPIPS, PSNR, SSIM\\
        \textit{Distributional:} R3D-Fr., JEDi\\
        \textit{Semantic:} CLIPScore, DreamSim\\[5pt]
        \textbf{Statistical tests}\\
        paired $t$, Wilcoxon signed-rank,\\
        win-rate, $p$-values};

    \draw[arrow] (ego.east) -- (egocorp.west);
    \draw[arrow] (exo.east) -- (exocorp.west);

    \draw[arrow] (egocorp.east) -- (cfg1.west);
    \draw[arrow] (exocorp.east) -- (cfg2.west);
    \draw[arrow] (egocorp.east) -- ($(cfg3.west)+(0,0.18)$);
    \draw[arrow] (exocorp.east) -- ($(cfg3.west)+(0,-0.18)$);

    \draw[arrow] (cfg1.east) -- ($(model.west)+(0,0.85)$);
    \draw[arrow] (cfg3.east) -- (model.west);
    \draw[arrow] (cfg2.east) -- ($(model.west)+(0,-0.85)$);

    \draw[arrowstrong] (model.east) -- (evalp.west);

    \end{tikzpicture}%
    }
    \caption{\textbf{Pipeline overview.} The RetailSMV corpus (\cref{sec:dataset}) provides synchronized egocentric and exocentric video across five real-world supermarkets, yielding $16{,}120$ ego and $15{,}985$ exo clips. Three matched LoRA configurations of the pretrained Cosmos3-Nano foundation video model~\cite{cosmos_paper} (\emph{egocentric-only}, \emph{exocentric-only}, and \emph{combined}) are trained under identical hyperparameters and optimization budget, isolating training-data viewpoint as the variable of interest. Every configuration is evaluated on the same $200$-clip stratified held-out test set under a paired statistical protocol.}
    \label{fig:overview}
\end{figure*}
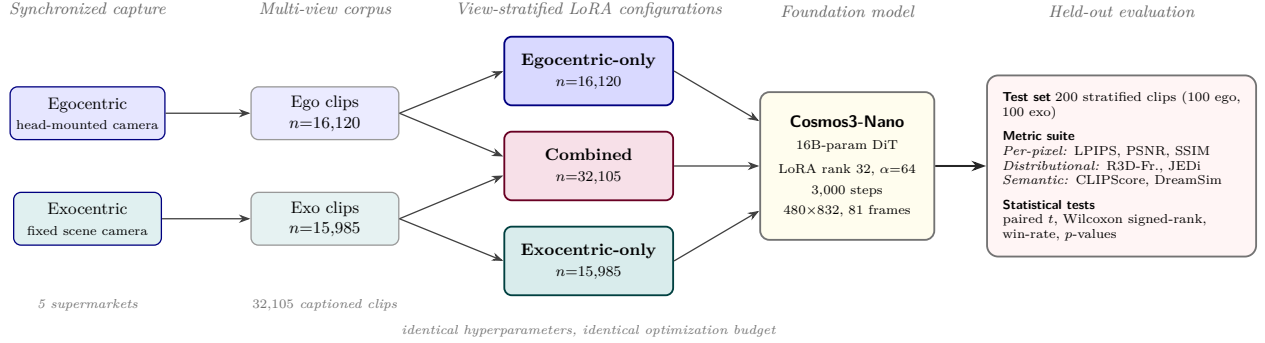

We organize the study around a single experimental question (see \cref{fig:overview} for the full pipeline):

\begin{center}\emph{Which viewpoint of training data (egocentric, exocentric, or their combination) produces the best adapted video world model for retail scenes?}\end{center}

To answer this we train three LoRA configurations of Cosmos3-Nano under matched hyperparameters and identical optimization budget: (i)~egocentric-only ($n{=}16{,}120$~clips), (ii)~exocentric-only ($n{=}15{,}985$~clips), and (iii)~combined ego$+$exo ($n{=}32{,}105$~clips). To control for seed variance we additionally train a second seed of the combined configuration. We evaluate every configuration on the same held-out test set of $200$ stratified retail clips balanced across both views ($n{=}100$ egocentric and $n{=}100$ exocentric), generating one video per (configuration, clip) pair under identical inference settings. The balanced stratification reflects both natural deployment viewpoints (the head-mounted egocentric camera and the fixed exocentric scene camera) at equal weight, and the egocentric subset enables a cross-view transfer check (\cref{sec:appendix_egoview}). The test set is disjoint from both the training data and the adapter-selection validation pool, eliminating selection leakage (\cref{sec:eval_setup}).

\nbf{Findings} Our analysis surfaces three robust empirical findings:

\begin{enumerate}[leftmargin=*,nosep]
    \item \emph{Adaptation succeeds uniformly.} All three LoRA configurations reduce the validation diffusion loss by a factor of approximately~$2.8$ (from $1.006$ to a tight range of $0.355{-}0.363$). Every one of $200$ paired evaluation samples improves over the pretrained baseline under each LoRA configuration, with both parametric and non-parametric paired tests at $p\ll 0.001$ (\cref{tab:val_loss}; exact values in \cref{sec:appendix_extra}). Seed-induced variation on the combined configuration is smaller than the configuration-to-configuration spread, supporting the interpretation that configuration rankings reflect properties of the training data rather than optimization noise.

    \item \emph{Exocentric-only adaptation matches or exceeds combined adaptation on six of seven point estimates and is significantly better on LPIPS, PSNR, and DreamSim.} Despite training on only the $15{,}985$ exocentric clips (versus the $32{,}105$ clips available to the combined configuration), the exocentric-only adapter achieves the lowest R3D-Fréchet ($35.14$ vs.\ base $42.69$, a $17.7\%$ relative reduction; combined $36.20$, a $15.2\%$ reduction), the lowest JEDi ($0.829$ vs.\ base $1.246$, a $33.5\%$ relative reduction; combined $0.892$, a $28.5\%$ reduction), the largest LPIPS reduction of $0.058$ at $100\%$ paired win-rate ($p{<}10^{-34}$), the largest PSNR improvement of $+0.575$ dB at $85\%$ win-rate, and the largest DreamSim reduction of $0.062$ at $92.5\%$ win-rate ($p{<}10^{-31}$). The combined configuration matches exo on SSIM ($+0.022$) and CLIPScore ($+0.017$ in the Hessel formulation) but is significantly \emph{worse} than exo on the perceptual and pixel-fidelity metrics in direct paired tests (LPIPS $p{<}10^{-7}$, PSNR $p{=}0.003$, DreamSim $p{<}10^{-9}$). The two distributional metrics agree on the ranking exo $<$ combined $<$ ego $<$ base, so the conclusion does not rely on either single feature backbone. This result suggests that combined ego$+$exo training does not improve over exo-only training on this corpus, and that the in-distribution exocentric subset is what carries the adaptation signal.

    \item \emph{The adaptation gap is concentrated in the near-horizon prediction window.} The absolute LPIPS gap to the pretrained base is largest at the shortest rollout time we sample ($t{=}0.5$ s, base mean LPIPS $0.542$ vs.\ exocentric-only $0.460$, a gap of $0.082$) and narrows steadily as $t$ grows. Restricting the LPIPS analysis to $t{=}1.0$ second of rollout widens the paired gap from the full-clip averages ($+0.041$, $+0.058$, $+0.049$ for ego, exo, combined) to $+0.061$, $+0.075$, and $+0.068$ respectively, at $82{-}89\%$ paired win-rate ($p{<}10^{-20}$). The near-horizon window is precisely the regime that world-model literature targets for embodied use~\cite{gaia2}.
\end{enumerate}

\nbf{Contributions} We make four contributions to the study of pretrained video world model adaptation:

\begin{enumerate}[leftmargin=*,nosep]
    \item We introduce \textbf{RetailSMV} (\emph{Retail Synchronized Multi-View}), a retail video corpus of $32{,}105$ captioned clips collected across five real-world supermarkets, with \emph{synchronized} egocentric and exocentric capture of \emph{store-staff} operational work (stocking, arranging, weighing, managing supply carts, scanning at checkout) rather than the customer-side behaviour captured by prior retail video corpora, dense paragraph-level captions, and pre-defined train, validation, and test splits. To our knowledge no prior dataset combines a real-world retail deployment domain with synchronized staff-perspective ego/exo capture of the same activity at this scale.

    \item Using RetailSMV, we present, to our knowledge, the first controlled comparison of egocentric-only, exocentric-only, and combined LoRA configurations of a foundation video world model, isolating training-data viewpoint as the variable of interest while holding all other factors constant.

    \item We document a reproducible \emph{headline} finding: exocentric-only adaptation matches or exceeds combined adaptation on six of seven point estimates and is significantly better on LPIPS, PSNR, and DreamSim, despite using only the $15{,}985$ exocentric training clips while combined uses the full $32{,}105$. The combined configuration edges exo on SSIM by a small margin, and the two are statistically indistinguishable on CLIPScore. We additionally characterize the temporal structure of the adaptation gap and show that the absolute LPIPS gap to the pretrained base is largest at the shortest rollout time and narrows steadily as $t$ grows.

    \item We provide a fully paired statistical protocol for video world model evaluation and report all metrics under both parametric ($t$-test) and non-parametric (Wilcoxon signed-rank) tests, addressing the documented absence of paired statistical reporting in current video-generation literature.
\end{enumerate}

The remainder of the paper is organized as follows. \Cref{sec:related} reviews related work on foundation video world models, video generation evaluation, and parameter-efficient adaptation. \Cref{sec:method} describes the multi-view retail data, the three view-stratified LoRA configurations, and the training and inference protocols. \Cref{sec:eval_setup} defines the test set, the metric suite, and the paired statistical tests. \Cref{sec:results} presents the main results, the temporal analysis, and metric ablations. \Cref{sec:conclusion} concludes with discussion and limitations.

\section{Related Work}
\label{sec:related}

\nbf{Foundation video generation models}
Text-to-video and image-to-video diffusion models have advanced rapidly. Make-A-Video~\cite{makeavideo} and AnimateDiff~\cite{animatediff} demonstrated that motion priors can be learned without paired text-video data; DynamiCrafter~\cite{dynamicrafter} and VideoCrafter2~\cite{videocrafter2} extended these to open-domain image-to-video conditioning; CogVideoX~\cite{cogvideox}, HunyuanVideo~\cite{hunyuanvideo}, Lumiere~\cite{lumiere}, Movie Gen~\cite{moviegen2024}, Open-Sora~\cite{opensora2024}, Stable Video Diffusion~\cite{svd2023}, Sora~\cite{sora2024}, and VideoPoet~\cite{videopoet} pushed scale, fidelity, and duration. The NVIDIA Cosmos family of world foundation models~\cite{cosmos_paper} targets embodied AI specifically and is the basis of our adaptation experiments. The underlying training objectives include denoising diffusion~\cite{ddpm,ddim}, flow matching~\cite{flowmatching}, and rectified flow~\cite{rectifiedflow}, with UniPC~\cite{unipc} now a common deterministic sampler at inference. Almost all of these models are pretrained on internet-scale generic video and report no domain-specific evaluation in retail environments.

\nbf{World models for embodied simulation}
The framing of pretrained video models as \emph{world simulators} for embodied agents traces to the world-models lineage of Ha and Schmidhuber~\cite{ha2018worldmodels} and is now exemplified by DreamerV3~\cite{dreamerv3}, Genie~\cite{genie}, and the learning-an-interactive-simulator framework of UniSim~\cite{unisim}, as well as more recent surveys~\cite{gao2025worldmodel}. Domain-specialized driving world models add explicit action conditioning: GAIA-1~\cite{gaia1}, GAIA-2~\cite{gaia2}, Vista~\cite{vista2024}, and DriveDreamer~\cite{drivedreamer}. Embodied robotics has produced action-conditioned generative policies that consume or emit video: RT-2~\cite{rt2}, PaLM-E~\cite{palme}, GR00T~\cite{groot_n1}, $\pi_0$~\cite{pi0}, AgiBot~\cite{agibot}, and the Gemini Robotics family~\cite{gemini_robotics}, supported by retail-adjacent benchmarks such as RoboVQA~\cite{robovqa}. Outside driving, however, systematic studies of adapting a foundation video model to a specific deployment environment are scarce.

\nbf{Parameter-efficient adaptation of diffusion models}
Adapting a large pretrained diffusion model to a new domain is typically done through parameter-efficient fine-tuning. The dominant family of methods comprises Low-Rank Adaptation~\cite{lora} and its variants~\cite{adalora,qlora}; subject-driven personalization with DreamBooth~\cite{dreambooth}; word-level personalization via textual inversion~\cite{textualinversion}; and multi-concept composition~\cite{customdiffusion}. For video specifically, motion-customization methods have started to appear: MotionDirector~\cite{motiondirector}, MotionCtrl~\cite{motionctrl}, and AnimateDiff~\cite{animatediff} all adopt LoRA-style adapters to inject motion or domain priors into a frozen image- or video-diffusion backbone. NVIDIA's Cosmos-Predict~2.5 cookbook~\cite{cosmos_paper} provides documented LoRA post-training recipes. Most prior video LoRA work treats viewpoint as a nuisance variable or focuses on a single viewpoint; to our knowledge there is no prior systematic comparison of egocentric, exocentric, and combined LoRA configurations on a synchronized multi-view video corpus.

\nbf{Video generation evaluation}
The standard suite is built around the Fréchet video distance (FVD)~\cite{fvd_original} on I3D Kinetics features~\cite{i3d}, per-frame LPIPS~\cite{lpips} with AlexNet features~\cite{clip_radford}, PSNR and SSIM~\cite{ssim}, and CLIPScore~\cite{clipscore} computed from CLIP image-text embeddings~\cite{clip_radford}. VBench~\cite{vbench} and VBench-2~\cite{vbench2} provide curated capability dimensions for generic video generation. These benchmarks were designed assuming large evaluation sets; FVD is documented to require large samples for stable covariance estimation~\cite{luo2024beyondfvd}, a regime rarely available in domain-specific applications. Modern alternatives respond to this gap. DreamSim~\cite{fu2023dreamsim} combines CLIP, OpenCLIP, and DINO~\cite{dino} features with human-similarity fine-tuning, producing a perceptual metric that outperforms LPIPS at semantic structure and object identity. JEDi~\cite{luo2024beyondfvd} replaces I3D-FVD with an MMD on V-JEPA features~\cite{vjepa}, reaching a stable estimate at far smaller sample counts. We adopt DreamSim and JEDi alongside the classical suite, use the Hessel formulation of CLIPScore~\cite{clipscore}, and use an R3D-Fréchet variant of the Kinetics-based Fréchet distance with the ResNet-3D-18 backbone~\cite{r3d18} described in \cref{sec:metrics}.

\nbf{Egocentric and exocentric video understanding}
Egocentric video research is dominated by Ego4D~\cite{ego4d}, Epic-Kitchens~\cite{epic_kitchens}, HOI4D~\cite{hoi4d}, and Aria~\cite{aria}, all of which skew toward kitchens, outdoor activities, and daily living scenes. EgoSchema~\cite{egoschema} provides a long-form egocentric video understanding benchmark. Charades-Ego~\cite{charades_ego} and Ego-Exo4D~\cite{egoexo4d} introduce paired ego/exo recordings and demonstrate that joint training improves cross-view action recognition and retrieval. The pretext-task literature has used arrow-of-time prediction~\cite{arrow_of_time} and spatio-temporal jigsaw solving~\cite{jigsaw} to learn temporal structure from unlabelled video. These studies focus on \emph{recognition} and \emph{representation learning}; we focus on \emph{generation}, where the question is qualitatively different: does adding ego data to exo data improve generation of new retail scenes, or does it dilute the learning signal? Our experiments show that on retail video the empirical answer is that combined ego$+$exo training does not improve distributional or semantic metrics over exo-only training: exocentric-only adaptation matches or beats combined adaptation on R3D-Fréchet and DreamSim. \Cref{tab:cosmos_comparison} summarizes the gap in publicly available foundation video models with respect to retail coverage and synchronized multi-view data.

\begin{table}[t]
\centering
\small
\caption{Domain coverage of recent foundation video and world models. Existing models are pretrained on entertainment video, driving footage, or generic web video, with no documented retail-specific training or evaluation and no documented use of synchronized egocentric/exocentric capture of the same activity in their disclosed pretraining recipes. Pretraining corpora are not always fully disclosed; the table reflects what is documented in each model's public technical report. We adapt Cosmos3-Nano on \textbf{RetailSMV} (\cref{sec:dataset}), which provides $32{,}105$ captioned clips with synchronized paired views, enabling the controlled view-stratified study reported in this paper.}
\label{tab:cosmos_comparison}
\resizebox{\linewidth}{!}{%
\begin{tabular}{lllcc}
\toprule
\textbf{Model} & \textbf{Type} & \textbf{Pretraining domain} & \textbf{Retail} & \textbf{Ego$+$Exo pairs} \\
\midrule
Sora~\cite{sora2024}              & Text-to-video diffusion   & Internet video, entertainment      & \texttimes & \texttimes \\
Stable Video Diffusion~\cite{svd2023} & Image-to-video diffusion & LAION$+$curated video          & \texttimes & \texttimes \\
Open-Sora~\cite{opensora2024}     & Text-to-video diffusion   & Open web video                     & \texttimes & \texttimes \\
Movie Gen~\cite{moviegen2024}     & Text-to-video diffusion   & Licensed entertainment video       & \texttimes & \texttimes \\
GAIA-1~\cite{gaia1}               & Action-conditioned WM    & 4{,}700 h driving                  & \texttimes & \texttimes \\
GAIA-2~\cite{gaia2}               & Action-conditioned WM, MV & 16{,}000 h driving, multi-camera   & \texttimes & \texttimes \\
Vista~\cite{vista2024}            & Action-conditioned WM    & nuScenes, Waymo                    & \texttimes & \texttimes \\
Cosmos3-Nano~\cite{cosmos_paper}  & Foundation video DiT (16B) & Internet$+$robot manipulation       & \texttimes & \texttimes \\
Cosmos-Predict 2.5~\cite{cosmos_paper} & Foundation video DiT (2B/14B) & Internet$+$robot manipulation     & \texttimes & \texttimes \\
\midrule
This work, RetailSMV (\cref{sec:dataset}) & \emph{(corpus)} & 5 supermarkets, real store-staff operations  & \checkmark & \checkmark \\
\bottomrule
\end{tabular}%
}
\end{table}

\nbf{Retail-specific video datasets and benchmarks}
Retail and shopping environments have been studied as a deployment domain across recognition, simulation, and language-grounded tasks. MERL Shopping~\cite{merl_shopping} provides surveillance-style exocentric clips of retail customer interactions; the retail-action benchmark~\cite{retailaction} and retail-vision~\cite{retailvision} survey extend this to fine-grained action recognition; SariBench~\cite{saribench} provides a simulated retail environment for embodied agents; PRISM~\cite{rouhi2026prism} is a recent multi-view multi-capability retail video dataset built for embodied vision-language models, with chain-of-thought supervision and an SFT-ready format. These works focus on \emph{recognition, question answering, or simulation}, not on \emph{video generation}; in particular none of them has been used to fine-tune or evaluate a foundation video diffusion model. They also focus on the \emph{customer side} of the retail scene (browsing, picking, paying). RetailSMV (this paper) is, to our knowledge, the first retail video corpus with synchronized ego/exo capture of the \emph{store-staff} operational viewpoint---stocking, arranging, weighing, managing supply carts, and scanning at the checkout---at a scale that supports finetuning of a foundation video world model (\cref{tab:dataset_comparison}). RetailSMV is distinct from PRISM: PRISM targets embodied vision-language supervision, whereas RetailSMV is organized for video-world-model adaptation with paired ego/exo clips, generation captions, adaptation splits, and generated-video evaluation protocols.

\nbf{Statistical reporting in video generation}
A practical observation in recent video-generation literature~\cite{vbench,gaia2,fu2023dreamsim} is that paired statistical tests, win-rates, and confidence intervals are rarely reported, even though metric standard deviations are often comparable in magnitude to the differences between methods. The general statistical-comparison framework of Dem{\v s}ar~\cite{demsar2006} predates the diffusion-model era but its core advice (report paired tests with both parametric and non-parametric variants) remains directly applicable. We adopt this protocol throughout: every per-clip metric is reported with mean improvement, paired win-rate, and both parametric ($t$-test) and non-parametric (Wilcoxon signed-rank) $p$-values.

\section{Method}
\label{sec:method}

We describe the RetailSMV corpus introduced with this paper (\cref{sec:dataset}), the parameter-efficient adaptation recipe applied to the pretrained foundation model (\cref{sec:lora}), the three view-stratified configurations whose comparison forms the core of our study (\cref{sec:configs}), and the inference protocol used to generate evaluation videos (\cref{sec:generation}).

\subsection{The RetailSMV Corpus}
\label{sec:dataset}

We introduce \textbf{RetailSMV} (\emph{Retail Synchronized Multi-View}), a retail video corpus of 32{,}105 captioned clips with synchronized egocentric and exocentric capture of the same store-staff activities across five real-world supermarkets. To our knowledge no prior dataset combines a real-world retail deployment domain with \emph{synchronized} ego/exo capture of the same activity at the scale required to fine-tune a foundation video world model. \Cref{tab:dataset_comparison} positions RetailSMV against related embodied video datasets.

\nbf{Capture protocol} The corpus is collected in five operating supermarkets that span distinct store layouts, lighting conditions, aisle configurations, and product category distributions. \textbf{Unlike prior retail video datasets which focus on customer-side behaviour}, RetailSMV is captured from the perspective of store staff performing their day-to-day operational work: stocking shelves, arranging produce, weighing and labelling items, carrying crates, pushing supply carts, measuring shelf space, scanning items at the checkout, and assisting customers. Each work episode is recorded simultaneously by two complementary sensors. The \textbf{egocentric} sensor is a head-mounted camera worn by the staff member while they carry out these tasks. The \textbf{exocentric} sensor is a fixed scene-level camera that observes the same activity from a stable third-person perspective. Synchronized capture means that for every ego clip there exists an exo clip of the identical activity at the same wall-clock time, with the same person, products, and scene context, captured from the complementary viewpoint. This pairing is what enables the controlled view ablations reported in this paper.

\nbf{Clip-level selection and captioning} From the raw recordings we extract 32{,}105 short clips ($1$--$8$\,s each) that satisfy three conditions: (i) the clip captures a complete atomic store-staff action, (ii) the synchronized view from the paired sensor is available for the same time interval, and (iii) a dense paragraph-level caption is available. Each caption is a single paragraph describing the visible store and section, the person's high-level task and low-level action, hand states and body pose, visible products and signage, and a frame-by-frame motion summary. The same caption is used as the text-conditioning input for every configuration of every training run, eliminating prompt variance as a source of cross-configuration difference. The full clip pool decomposes into $16{,}120$ egocentric and $15{,}985$ exocentric clips, an essentially balanced split that lets us match clip count exactly across single-view configurations.

\nbf{Privacy} All faces are de-identified through Gaussian blurring at the source. This is applied uniformly to both ego and exo views and to the train, validation, and test splits; we do not introduce or remove any face obfuscation in this work.

\begin{table*}[t]
\centering
\small
\caption{\textbf{Positioning of RetailSMV against related embodied video datasets.}
RetailSMV is, to our knowledge, the only embodied video dataset that combines a real-world retail deployment domain, \emph{synchronized} egocentric and exocentric capture of the same activity, and dense paragraph-level captioning at a scale that supports finetuning of a foundation video diffusion model.
``Sync ego$+$exo'' indicates that ego and exo recordings of the same activity at the same wall-clock time are available and identified at the clip level (not merely that both modalities exist in the corpus).
``Paragraph caption'' indicates a single dense natural-language paragraph per clip suitable as text conditioning for a text-and-image-to-video model.
$^{\dagger}$Ego-Exo4D pairs ego and exo views but is concentrated on skilled-activity domains (sports, music, cooking) rather than retail.}
\label{tab:dataset_comparison}
\resizebox{\linewidth}{!}{%
\begin{tabular}{lllcccr}
\toprule
\textbf{Dataset} & \textbf{Domain} & \textbf{Viewpoints} & \textbf{Sync ego$+$exo} & \textbf{Retail} & \textbf{Paragraph caption} & \textbf{Scale (clips/hours)} \\
\midrule
Ego4D~\cite{ego4d}                  & Daily living, general              & Ego                          & \texttimes      & \texttimes & \texttimes & 3{,}670\,h \\
Ego-Exo4D~\cite{egoexo4d}           & Skilled activities$^{\dagger}$     & Ego $+$ Exo                  & \checkmark      & \texttimes & \texttimes & 1{,}286\,h \\
Charades-Ego~\cite{charades_ego}    & Daily living                       & Ego $+$ Exo                  & \texttimes      & \texttimes & \texttimes & 68.8\,h    \\
Epic-Kitchens~\cite{epic_kitchens}  & Kitchen                            & Ego                          & \texttimes      & \texttimes & \texttimes & 100\,h     \\
Aria Everyday Activities~\cite{aria}& Daily living                       & Ego                          & \texttimes      & \texttimes & \texttimes & ---        \\
RoboVQA~\cite{robovqa}              & Office robot manipulation          & Exo                          & \texttimes      & \texttimes & \texttimes & 829{,}000 pairs \\
SariBench~\cite{saribench}          & Retail (simulator)                 & Ego (simulated)              & \texttimes      & \checkmark & \texttimes & 100 demos  \\
\midrule
\textbf{RetailSMV (ours)}            & \textbf{Real retail (5 supermarkets)} & \textbf{Ego $+$ fixed Exo}         & \textbf{\checkmark}      & \textbf{\checkmark}  & \textbf{\checkmark}  & \textbf{32{,}105 clips} \\
\bottomrule
\end{tabular}%
}
\end{table*}

\nbf{Train, validation, and test splits} We partition RetailSMV into three pairwise disjoint splits by per-clip unique identifier. The training set consists of 32{,}105 clips and is used for LoRA adaptation. The validation set consists of 1{,}388 clips and is used for adapter selection: the rectified-flow validation loss is computed every 100 training steps on $n{=}32$ paired samples drawn from it. The test set consists of 200 clips drawn via stratified sampling from the held-out evaluation pool, balanced across both egocentric ($n{=}100$) and exocentric ($n{=}100$) viewpoints so that both deployment scenarios are represented at equal weight in the per-clip paired statistics. The egocentric-only model is evaluated on its in-distribution view as well as on the exocentric view, exposing cross-view transfer behavior. \Cref{tab:dataset_overview} summarizes the splits.

\begin{table}[t]
\centering
\small
\caption{Splits used in this study. Training, validation (used for adapter selection), and test (used for final evaluation) are pairwise disjoint by per-clip unique identifier.}
\label{tab:dataset_overview}
\begin{tabular}{lrrr}
\toprule
\textbf{Split} & \textbf{Ego} & \textbf{Exo} & \textbf{Total} \\
\midrule
Training & 16{,}120 & 15{,}985 & \textbf{32{,}105} \\
Validation (adapter selection) & 727 & 661 & 1{,}388 \\
Test (final evaluation) & 100 & 100 & \textbf{200} \\
\bottomrule
\end{tabular}
\end{table}

\subsection{LoRA Adaptation of the Foundation Model}
\label{sec:lora}

We adapt the open-source Cosmos3-Nano video diffusion model~\cite{cosmos_paper}, a 16-billion-parameter Mixture-of-Transformers Diffusion Transformer with a Wan2.2 video VAE and the Cosmos-Reason~\cite{cosmos_reason1,cosmos_reason2} text encoder, using Low-Rank Adaptation~\cite{lora}.

\nbf{LoRA configuration} We attach LoRA adapters~\cite{lora} of rank $r{=}32$ and scaling $\alpha{=}64$ to the cross-attention projections and the Mixture-of-Experts MLPs in every generative-modality transformer block, training $\sim 0.55\%$ of the base model's parameters. This follows the standard adapter-pluggable convention now widely used for diffusion-model personalization~\cite{dreambooth,textualinversion,customdiffusion,adalora} and motion customization~\cite{motiondirector,motionctrl,animatediff}. Standard zero-initialization of $B$ ensures the model at step zero reproduces the pretrained base. Full layer-wise target list, exact parameter count, and AdamW~\cite{adamw} settings are in \cref{tab:hp}.

\nbf{Training objective} We train under the rectified-flow~\cite{rectifiedflow} velocity-matching objective used by Cosmos3-Nano during its own pretraining, a stochastic-interpolant family closely related to denoising diffusion~\cite{ddpm,ddim} and flow matching~\cite{flowmatching}. For each training step we encode the input clip to its latent representation $z_0$, draw a noise scale $\sigma \sim \mathrm{Sigmoid}(\mathcal{N}(0,1))$, sample a Gaussian noise tensor $\epsilon\sim\mathcal{N}(0,\mathbf I)$, form the perturbed latent $z_\sigma = \sigma\epsilon + (1{-}\sigma)z_0$, and target the velocity $v_\sigma = \epsilon - z_0$. The training loss is the mean squared error $\mathcal{L} = \|\hat v_\sigma - v_\sigma\|_2^2$ averaged over the spatio-temporal latent grid.

\nbf{Training protocol} We train at $480{\times}832$ spatial resolution, $81$ frames per clip, for $3{,}000$ optimization steps under a cosine learning-rate schedule with a peak rate of $3{\times}10^{-4}$. Validation diffusion loss is computed every $100$ steps on $n{=}32$ paired samples drawn from the $1{,}388$-clip validation set, and we select the iteration-$1500$ adapter (lowest validation loss) for all subsequent generation experiments. The full training environment (precision, video decoder, hardware, effective batch size, and other infrastructure details) is in \cref{sec:appendix_hp,tab:hp}.

\subsection{View-Stratified Configurations}
\label{sec:configs}

To isolate the effect of training-data viewpoint while holding all other variables constant, we train three matched configurations under identical hyperparameters and optimization budget:

\nbf{Combined (ego $+$ exo)} The reference configuration, trained on the full 32{,}105-clip mixture. This corresponds to the natural default of using all available data.

\nbf{Egocentric-only} Trained on the 16{,}120 egocentric clips alone. This configuration can specialize to first-person motion patterns and hand-object interaction, at the cost of losing exposure to wide-field scene structure.

\nbf{Exocentric-only} Trained on the 15{,}985 exocentric clips alone. This configuration can specialize to scene layout, multi-actor dynamics, and the wide field of view of the fixed scene camera, at the cost of losing exposure to fine-grained hand interaction.

\nbf{Combined, second seed} An additional combined configuration trained with a different random seed (137 instead of 42), used to bound the seed-induced variance on the validation metrics and to confirm that configuration-to-configuration differences exceed this variance.

\subsection{Inference Protocol}
\label{sec:generation}

For evaluation we generate one video per (configuration, test clip) pair under image-to-video conditioning: the first frame of the ground-truth test clip serves as the conditioning image, and the dense caption for that clip serves as the text prompt. We generate $189$ frames at $480{\times}832$, the native maximum length of Cosmos3-Nano ($\sim 7.9$\,s at $24$~FPS), so that wall-clock time-aligned metric sampling is well defined for the full duration of every clip. The deterministic UniPC multistep sampler~\cite{unipc} is used at inference. Scheduler, inference steps, guidance scale, negative prompt, and seed are listed in \cref{sec:appendix_gen}. The trained adapter is merged into the base weights at inference; we report a numerical merge verification in \cref{sec:appendix_merge}.

\section{Evaluation Protocol}
\label{sec:eval_setup}

This section defines the test set (\cref{sec:eval_set}), the metric suite (\cref{sec:metrics}), the paired statistical protocol (\cref{sec:stats}), and the near-horizon analysis used to characterize the temporal structure of the adaptation gap (\cref{sec:restrictions}).

\subsection{Test Set}
\label{sec:eval_set}

Due to the high inference cost of video generation (on the order of minutes per clip on contemporary hardware), we evaluate on a small held-out test set, the standard mode of evaluation in current video-generation literature. We use the same test set across all four configurations and the pretrained baseline, generated under identical inference settings (\cref{sec:generation}).

The test set consists of $200$ retail clips drawn from a held-out evaluation pool via stratified sampling, balanced across both egocentric ($n{=}100$) and exocentric ($n{=}100$) viewpoints. The draw is deterministic and uses rejection sampling against the validation set used for adapter selection. We verify by per-clip unique-identifier intersection that the test set is disjoint from both the training data and the adapter-selection validation set.

\nbf{Asset pipeline} For each test clip we extract the full-length ground-truth video, the first frame as a JPEG image, and the dense caption. We then generate exactly one video per (configuration, test clip) pair, for a total of $4{\times}200=800$ generated videos. All generations use the deterministic settings of \cref{sec:generation}.

\subsection{Metric Suite}
\label{sec:metrics}

We report a deliberately wide panel of seven metrics to avoid over-interpreting any single number. All metrics are computed on the same generated videos. The panel comprises per-frame perceptual metrics (LPIPS, PSNR, SSIM), two distributional metrics (a classical Kinetics-feature Fréchet distance and a recent V-JEPA-based MMD), a semantic alignment metric (Hessel CLIPScore), and a modern human-aligned perceptual metric (DreamSim). We report JEDi as a recent distributional video-quality metric \emph{complementary} to R3D-Fr\'echet; we do not rely on JEDi alone, and the same ranking is supported by R3D-Fr\'echet and DreamSim.

\nbf{LPIPS}~\cite{lpips} measures perceptual distance between matched generated/ground-truth frame pairs using AlexNet features. We sample 16 evenly-spaced wall-clock timestamps in $[0,\min(\text{gen\_dur},\text{gt\_dur})]$, dropping $t{=}0$ to remove the trivial first-frame match introduced by image-to-video conditioning. Lower is better. The rationale for wall-clock sampling (avoiding a frame-index pitfall between $24$~FPS generations and lower-FPS ground truths) is detailed in \cref{sec:appendix_time_align}.

\nbf{PSNR and SSIM}~\cite{ssim} are computed per frame on the same time-aligned samples used for LPIPS and averaged over the clip. Higher is better.

\nbf{Hessel CLIPScore}~\cite{clipscore} measures caption-video alignment. For each of 8 evenly-spaced frames of the generated video, we compute $2.5\cdot\max(0,\cos(\text{img\_emb},\text{txt\_emb}))$ with CLIP ViT-B/32 (Hessel et al.'s convention), and average across frames. Higher is better; typical values fall in the $0.6$--$0.85$ range.

\nbf{R3D-Fréchet video distance} We compute a Fréchet distance on features from the ResNet-3D-18 video classifier (\texttt{torchvision.models.video.r3d\_18} with \texttt{KINETICS400\_V1} weights) on 16 frames per clip, used in place of the canonical I3D~\cite{fvd_original,i3d}. We call the resulting metric \emph{R3D-Fréchet} to make the backbone substitution explicit; absolute values are not directly comparable to I3D-FVD values in the broader literature. Lower is better. See \cref{sec:appendix_notes} for the rationale and a recipe for recomputing with the canonical I3D backbone.

\nbf{DreamSim}~\cite{fu2023dreamsim} is a perceptual distance built from an ensemble of CLIP, OpenCLIP, and DINO features fine-tuned on human similarity triplets. It is documented to outperform LPIPS, raw CLIP, and DINO on human similarity benchmarks. We compute it on the same 16 time-aligned frames used for LPIPS, averaged over the clip. Lower is better.

\nbf{JEDi}~\cite{luo2024beyondfvd} is a recent distributional video-quality metric that replaces the I3D-FVD covariance estimate with a maximum mean discrepancy on V-JEPA~\cite{vjepa} ViT-H/16 features. It is documented to reach a stable estimate at far smaller sample counts than I3D-FVD. We compute JEDi on $16$ uniformly-sampled frames per clip at $224{\times}224$, using the official \texttt{videojedi} package with the V-JEPA ViT-H/16 backbone, with the gold (ground-truth) loader and the generated loader containing the same set of clips. Lower is better. The reported JEDi value is a biased empirical polynomial-kernel MMD estimator and its absolute scale is sample-set dependent (we verified this directly by recomputing JEDi on disjoint sub-pools of our test set); the relative reduction vs.\ the pretrained base, and the configuration ranking, are the reportable quantities. We report JEDi as a \emph{complementary} distributional signal to R3D-Fréchet; we do not rely on it alone, and the configuration ranking is consistent with R3D-Fréchet and DreamSim.

\nbf{Validation diffusion loss} For each configuration we additionally compute the rectified-flow training loss on a paired-noise evaluation set of $n{=}200$ random samples from the validation set used for adapter selection (\emph{not} from the held-out test set). To ensure exact pairing across configurations, the noise tensor and noise scale for sample $i$ are derived from a fixed seed $10000{+}i$ shared by all configurations and the base. This produces per-sample loss values that are directly comparable across configurations and forms the basis of \cref{tab:val_loss}.

\subsection{Paired Statistical Protocol}
\label{sec:stats}

For every per-clip metric and every LoRA configuration $c$ we compute, for each test clip $i$, the signed paired difference
\begin{equation*}
\delta_i(c) \;=\;\begin{cases}m_i(\text{base})-m_i(c),& m\text{ lower-is-better}\\ m_i(c)-m_i(\text{base}),& m\text{ higher-is-better}\end{cases}
\end{equation*}
so that $\delta_i(c){>}0$ means $c$ beats the base on clip $i$. We then report (i) the mean paired difference $\bar\delta(c)$, (ii) the paired win-rate $|\{i:\delta_i(c){>}0\}|/n$, (iii) a parametric two-sided paired $t$-test $p$-value, and (iv) a non-parametric Wilcoxon signed-rank $p$-value. We treat $p{<}10^{-4}$ as highly significant. For the distributional R3D-Fréchet metric we report the per-configuration scalar score and the relative improvement, since it operates on the set of videos as a whole rather than on paired samples.

\subsection{Near-Horizon LPIPS Analysis}
\label{sec:restrictions}

World models are predominantly used to predict the immediate future of a scene, typically the next one to three seconds~\cite{gaia2}. The mean LPIPS curves (\cref{fig:lpips_vs_time}) rise monotonically with $t$ for every configuration: the absolute gap to the pretrained base is widest at the shortest rollout time we sample ($t{=}0.5$\,s) and narrows steadily as rollout time grows, since all configurations drift away from the single ground-truth trajectory in the same way. To characterize the temporal structure of the adaptation gap, we additionally report LPIPS at fixed time stamps $t\in\{1.0,2.0\}$ seconds on the same test clips. This is a reporting convention that we recommend for future video world model evaluations, since it isolates the regime in which the model's prediction quality matters most for downstream embodied use.

\section{Results and Analysis}
\label{sec:results}

We present the results in five subsections. \Cref{sec:res_valloss} establishes that all three view-stratified configurations improve uniformly over the pretrained baseline on the unified validation diffusion loss. \Cref{sec:res_main} reports the main paired metrics on the held-out test set, with direct paired comparisons between the adapted configurations (combined vs.\ exocentric-only and combined vs.\ egocentric-only) and the asymmetric data-contribution result. \Cref{sec:res_qualitative} walks through ten representative video examples to ground the metric-level differences in concrete failure modes. \Cref{sec:res_temporal} examines the temporal structure of the adaptation gap and identifies the near-horizon prediction window as the regime in which the LoRA-adapted models differ most from the pretrained base. \Cref{sec:ablations} reports time-aligned-vs-index-based sampling, adapter-merge verification, and modern-metric sensitivity. All $p$-values are computed under the paired protocol of \cref{sec:stats}.

\subsection{Adaptation Improves Validation Loss Uniformly}
\label{sec:res_valloss}

\Cref{tab:val_loss} reports the rectified-flow validation loss on $200$ paired samples drawn from the validation set used for adapter selection. By construction, the noise tensor and noise scale for sample $i$ are identical across all configurations, so the per-sample loss values are directly comparable.

\begin{table}[t]
\centering
\small
\caption{Unified validation diffusion loss on $200$ paired samples drawn from the validation set used for adapter selection (no overlap with the held-out test set). The noise tensor and noise scale for sample $i$ are identical across all rows, so the per-sample values are directly comparable. ``Win'' is the paired win-rate over the pretrained base.}
\label{tab:val_loss}
\begin{tabular}{lcccc}
\toprule
\textbf{Configuration} & \textbf{Loss} $\downarrow$ & \textbf{$\Delta$ vs.\ base} & \textbf{Win} & \textbf{$p$ (Wilcoxon)} \\
\midrule
Pretrained base & $1.006\pm0.197$ & --- & --- & --- \\
Egocentric-only & $0.359\pm0.144$ & $-0.647$ & 100\% & $\ll 10^{-4}$ \\
Exocentric-only & $0.363\pm0.150$ & $-0.643$ & 100\% & $\ll 10^{-4}$ \\
Combined ($s{=}42$) & $0.355\pm0.145$ & $-0.651$ & 100\% & $\ll 10^{-4}$ \\
Combined ($s{=}137$) & $0.357\pm0.145$ & $-0.649$ & 100\% & $\ll 10^{-4}$ \\
\bottomrule
\end{tabular}
\end{table}

Three observations follow. First, adaptation reduces validation loss by a factor of approximately $2.8$, and every one of the $200$ paired evaluation samples improves over the pretrained base under each configuration (perfect paired ordering under the Wilcoxon signed-rank test; exact $p$-values are reported in \cref{sec:appendix_extra}). Second, the seed-induced variation between the two combined seeds is $0.002$, smaller than the configuration-to-configuration spread of $0.008$ across the four LoRA rows; configuration-to-configuration ordering therefore reflects properties of the training data rather than optimization noise. Third, the three view-stratified configurations are within $0.008$ of each other on this metric, which means validation loss alone does not discriminate between view-stratified data choices. Differences across configurations emerge when generation quality is evaluated directly.

\subsection{Main Paired Metrics}
\label{sec:res_main}

\Cref{tab:main} reports the seven primary metrics on the held-out 200-clip test set. The exocentric-only configuration achieves the lowest R3D-Fréchet ($35.14$ vs.\ base $42.69$, a $17.7\%$ relative reduction), the lowest JEDi ($0.829$ vs.\ base $1.246$, a $33.5\%$ relative reduction), the largest LPIPS reduction of $0.058$ at $100\%$ paired win-rate ($p{<}10^{-34}$), the largest PSNR improvement of $+0.575$\,dB at $85\%$ paired win-rate, and the largest DreamSim reduction of $0.062$ at $92.5\%$ paired win-rate ($p{<}10^{-31}$). The combined configuration is tied with exo on CLIPScore and edges exo on SSIM by $0.0012$. Hessel CLIPScore improves under all three configurations with very strong paired significance ($p{<}10^{-10}$). We report JEDi as a recent distributional video-quality metric \emph{complementary} to R3D-Fr\'echet: we do not rely on JEDi alone, and the same ranking is supported by R3D-Fr\'echet and DreamSim.

\begin{table*}[t]
\centering
\caption{Main results on the held-out 200-clip test set. For lower-is-better metrics (LPIPS, DreamSim, R3D-Fréchet, JEDi) the best value in each column is in \textbf{bold}; for higher-is-better metrics (PSNR, SSIM, CLIPScore) the largest value is in \textbf{bold}. CLIPScore is in the Hessel $2.5\cdot\max(0,\cos)$ formulation. Paired $p$-values are Wilcoxon signed-rank against the pretrained base. R3D-Fréchet and JEDi are set-level distributional metrics and admit no paired test; their relative improvement vs.\ base is reported parenthetically.}
\label{tab:main}
\resizebox{\textwidth}{!}{%
\begin{tabular}{lccccccc}
\toprule
\textbf{Configuration} & \textbf{LPIPS} $\downarrow$ & \textbf{PSNR} $\uparrow$ & \textbf{SSIM} $\uparrow$ & \textbf{CLIPScore} $\uparrow$ & \textbf{DreamSim} $\downarrow$ & \textbf{R3D-Fr.} $\downarrow$ & \textbf{JEDi} $\downarrow$ \\
\midrule
\multicolumn{8}{l}{\textit{Per-clip mean (where applicable)}} \\
Pretrained base & $0.668$ & $10.68$ & $0.215$ & $0.815$ & $0.263$ & $42.69$ & $1.246$ \\
Egocentric-only & $0.626$ & $11.01$ & $0.226$ & $0.832$ & $0.220$ & $37.65$ ($-11.8\%$) & $0.960$ ($-22.9\%$) \\
Exocentric-only & $\mathbf{0.610}$ & $\mathbf{11.26}$ & $0.235$ & $\mathbf{0.834}$ & $\mathbf{0.201}$ & $\mathbf{35.14}$ ($\mathbf{-17.7\%}$) & $\mathbf{0.829}$ ($\mathbf{-33.5\%}$) \\
Combined & $0.619$ & $11.18$ & $\mathbf{0.236}$ & $0.832$ & $0.216$ & $36.20$ ($-15.2\%$) & $0.892$ ($-28.5\%$) \\
\midrule
\multicolumn{8}{l}{\textit{Paired vs.\ base: mean improvement / win-rate / Wilcoxon $p$}}\\
Egocentric-only & $+0.041$/$90.5\%$/${<}10^{-27}$ & $+0.33$/$72.0\%$/${<}10^{-12}$ & $+0.011$/$67.0\%$/${<}10^{-8}$ & $+0.017$/$72.5\%$/${<}10^{-10}$ & $+0.043$/$78.0\%$/${<}10^{-18}$ & --- & --- \\
Exocentric-only & $\mathbf{+0.058}$/$\mathbf{100\%}$/${<}\mathbf{10^{-34}}$ & $\mathbf{+0.58}$/$\mathbf{85.0\%}$/${<}\mathbf{10^{-25}}$ & $+0.021$/$76.0\%$/${<}10^{-17}$ & $\mathbf{+0.018}$/$\mathbf{75.0\%}$/${<}\mathbf{10^{-13}}$ & $\mathbf{+0.062}$/$\mathbf{92.5\%}$/${<}\mathbf{10^{-31}}$ & --- & --- \\
Combined & $+0.049$/$94.5\%$/${<}10^{-31}$ & $+0.50$/$78.5\%$/${<}10^{-18}$ & $\mathbf{+0.022}$/$\mathbf{80.0\%}$/${<}\mathbf{10^{-20}}$ & $+0.017$/$71.0\%$/${<}10^{-10}$ & $+0.046$/$82.0\%$/${<}10^{-23}$ & --- & --- \\
\bottomrule
\end{tabular}%
}
\end{table*}

\nbf{Headline finding: exocentric-only adaptation matches or exceeds combined adaptation on six of seven point estimates and is significantly better on LPIPS, PSNR, and DreamSim} The exocentric-only adapter, trained on $15{,}985$ exocentric clips, achieves an R3D-Fréchet of $35.14$, a $17.7\%$ relative reduction from the base, and a JEDi of $0.829$, a $33.5\%$ relative reduction (the largest distributional gap of any configuration); the combined adapter, with access to all $32{,}105$ clips, achieves $36.20$ R3D-Fr.\ ($-15.2\%$) and $0.892$ JEDi ($-28.5\%$). On LPIPS the exocentric-only adapter produces the largest paired improvement of $+0.058$ at $100\%$ paired win-rate (every one of $200$ test clips); combined is at $+0.049$ at $94.5\%$. On PSNR exo improves by $+0.58$\,dB at $85.0\%$ win-rate versus combined's $+0.50$\,dB at $78.5\%$. On DreamSim exo achieves a reduction of $0.062$ at $92.5\%$ win-rate ($p{<}10^{-31}$) versus combined's reduction of $0.046$ at $82.0\%$. The exocentric-only point estimate matches or exceeds the combined point estimate on six of the seven metrics we report (LPIPS, PSNR, CLIPScore, DreamSim, R3D-Fréchet, and JEDi); combined edges exo on SSIM only, by $0.0012$. Of the five per-clip metrics for which a paired test is defined, exo is significantly better than combined on LPIPS, PSNR, and DreamSim (\cref{tab:com_vs_exo}), and the two configurations are statistically indistinguishable on SSIM and CLIPScore. The two distributional metrics agree on the ranking (exo $<$ combined $<$ ego $<$ base), supporting the conclusion under either feature backbone.

\nbf{Direct paired comparison: combined vs.\ exocentric-only}
The paired Wilcoxon tests in \cref{tab:main} are against the pretrained base. To support the headline claim that combined ego$+$exo training does \emph{not} improve over exo-only, we additionally compute paired tests directly between the two adapted configurations on the same $200$-clip test set. \Cref{tab:com_vs_exo} reports the mean paired difference (combined $-$ exocentric-only), $95\%$ bootstrap confidence interval over clips, combined win-rate, and Wilcoxon $p$-value for every per-clip metric. Exo is significantly better than combined on LPIPS, PSNR, and DreamSim. Combined and exo are statistically indistinguishable on SSIM and CLIPScore (the $95\%$ CI for SSIM straddles zero, and the CLIPScore CI is symmetric around zero). For the set-level R3D-Fréchet metric we cannot run a paired test (it operates on whole feature sets), but the point estimate is also in favor of exocentric-only ($35.14$ vs.\ combined $36.20$).

\begin{table}[t]
\centering
\small
\caption{Direct paired comparison between the combined and exocentric-only configurations on the $200$-clip test set. $\Delta$ is the per-clip mean of (combined $-$ exocentric-only); $95\%$ CI is by clip-level bootstrap with $B{=}5000$ replicates; the win-rate is the fraction of clips on which combined beats exo. Combined is significantly \emph{worse} than exo on LPIPS, PSNR, and DreamSim; the SSIM and CLIPScore differences are not statistically distinguishable from zero.}
\label{tab:com_vs_exo}
\begin{tabular}{lrcrr}
\toprule
\textbf{Metric} & \textbf{$\Delta$} & \textbf{$95\%$ CI} & \textbf{Win-rate} & \textbf{Wilcoxon $p$} \\
\midrule
LPIPS $\downarrow$        & $+0.0086$ & $[+0.0054,\,+0.0118]$ & $33.0\%$ & ${<}10^{-7}$ \\
PSNR $\uparrow$            & $-0.075$  & $[-0.126,\,-0.025]$   & $40.0\%$ & $0.003$ \\
SSIM $\uparrow$            & $+0.0012$ & $[-0.0007,\,+0.0031]$ & $54.0\%$ & $0.18$ \\
CLIPScore $\uparrow$       & $-0.0015$ & $[-0.0049,\,+0.0019]$ & $44.5\%$ & $0.20$ \\
DreamSim $\downarrow$      & $+0.0154$ & $[+0.0105,\,+0.0204]$ & $32.0\%$ & ${<}10^{-9}$ \\
\bottomrule
\end{tabular}
\end{table}

\nbf{We find no statistically significant metric on which combined adaptation outperforms exocentric-only adaptation} The picture from \cref{tab:com_vs_exo} is consistent across metrics: on three of the five per-clip metrics (LPIPS, PSNR, DreamSim) combined is statistically \emph{worse} than exo at $p\leq 0.003$, with combined-wins fractions of $32$--$40\%$ (below chance), and the DreamSim gap is significant at $p{<}10^{-9}$. On SSIM and CLIPScore the two are statistically indistinguishable (the $95\%$ CI for SSIM straddles zero and the CLIPScore CI is symmetric around zero). The data-rich combined configuration does not deliver any statistically significant per-clip-metric advantage over the exo-only adapter; the in-distribution exocentric subset carries the adaptation signal.

\nbf{Direct paired comparison: combined vs.\ egocentric-only}
For a symmetric view of the asymmetry, we additionally compute paired tests directly between the combined and egocentric-only adapters on the same $200$-clip test set. \Cref{tab:com_vs_ego} reports the mean paired difference (combined $-$ egocentric-only), $95\%$ bootstrap confidence interval over clips, combined win-rate, and Wilcoxon $p$-value for every per-clip metric. Combined is significantly better than egocentric-only on LPIPS ($p{=}0.002$), PSNR ($p{<}10^{-6}$), and SSIM ($p{<}10^{-14}$); the two configurations are statistically indistinguishable on CLIPScore and DreamSim. For the set-level distributional metrics, combined is also in front of egocentric-only (R3D-Fr.\ $36.20$ vs.\ $37.65$; JEDi $0.892$ vs.\ $0.960$).

\begin{table}[t]
\centering
\small
\caption{Direct paired comparison between the combined and egocentric-only configurations on the $200$-clip test set. $\Delta$ is the per-clip mean of (combined $-$ egocentric-only); $95\%$ CI is by clip-level bootstrap with $B{=}5000$ replicates; the win-rate is the fraction of clips on which combined beats egocentric-only. Combined is significantly \emph{better} than egocentric-only on LPIPS, PSNR, and SSIM; the CLIPScore and DreamSim differences are not statistically distinguishable from zero.}
\label{tab:com_vs_ego}
\begin{tabular}{lrcrr}
\toprule
\textbf{Metric} & \textbf{$\Delta$} & \textbf{$95\%$ CI} & \textbf{Win-rate} & \textbf{Wilcoxon $p$} \\
\midrule
LPIPS $\downarrow$        & $-0.0075$ & $[-0.0122,\,-0.0032]$ & $60.0\%$ & $0.0018$ \\
PSNR $\uparrow$            & $+0.173$  & $[+0.103,\,+0.249]$   & $67.0\%$ & ${<}10^{-6}$ \\
SSIM $\uparrow$            & $+0.0105$ & $[+0.0078,\,+0.0134]$ & $77.5\%$ & ${<}10^{-14}$ \\
CLIPScore $\uparrow$       & $+0.0001$ & $[-0.0035,\,+0.0038]$ & $51.0\%$ & $0.83$ \\
DreamSim $\downarrow$      & $-0.0033$ & $[-0.0090,\,+0.0020]$ & $51.0\%$ & $0.53$ \\
\bottomrule
\end{tabular}
\end{table}

\nbf{Asymmetric data contribution: adding exocentric data to ego helps; adding egocentric data to exo hurts} Read alongside \cref{tab:com_vs_exo}, \cref{tab:com_vs_ego} exposes a clear asymmetry. Starting from the egocentric-only adapter, adding the exocentric subset (i.e., training combined) produces statistically significant improvements on LPIPS, PSNR, and SSIM at $p\leq 0.0018$ and improves the set-level distributional scores (R3D-Fréchet and JEDi). Starting from the exocentric-only adapter, however, adding the egocentric subset (again, training combined) produces statistically significant \emph{regressions} on LPIPS, PSNR, and DreamSim at $p\leq 0.003$. The two adapted configurations are statistically tied on CLIPScore in both directions. Combined therefore improves over egocentric-only on several metrics, but we find no statistically significant metric on which combined outperforms exocentric-only on this corpus, indicating that the adaptation signal is asymmetrically carried by the exocentric subset.

\nbf{Egocentric-only adaptation transfers but ranks third} The egocentric-only configuration improves over the base on every metric at $p{<}10^{-8}$, but it ranks behind both exo-only and combined on every column of \cref{tab:main}. The egocentric adapter does transfer non-trivially to the mixed test viewpoint (LPIPS paired win-rate is $90.5\%$ at $p{<}10^{-27}$) but the transfer is consistently weaker than the in-distribution exocentric-only adapter, indicating that paired ego$+$exo capture does not eliminate the need for view-matched training data.

\subsection{Qualitative Analysis}
\label{sec:res_qualitative}

The metric-level differences above translate cleanly into failure-mode patterns that are visible in the generated videos themselves. We walk through ten representative examples drawn from the held-out test set. Every figure in this subsection uses the same five-row layout: the top row is the ground-truth video (GT), and the four rows beneath show how the base, egocentric-only, exocentric-only, and combined configurations extend the shared conditioning frame over the next seven seconds. The column at $t{=}0$ is the shared conditioning frame seen by all four generated configurations, which is the same frame as the first GT frame. \emph{The first column ($t{=}0$) is therefore identical across all five rows by construction.} The cell highlighted in red marks, for each example, the base configuration's frame at which the failure is most visible (typically $t{=}1$, $t{=}3$, $t{=}5$, or $t{=}7$~s).

\begin{figure*}[t]
\centering
\includegraphics[width=\linewidth]{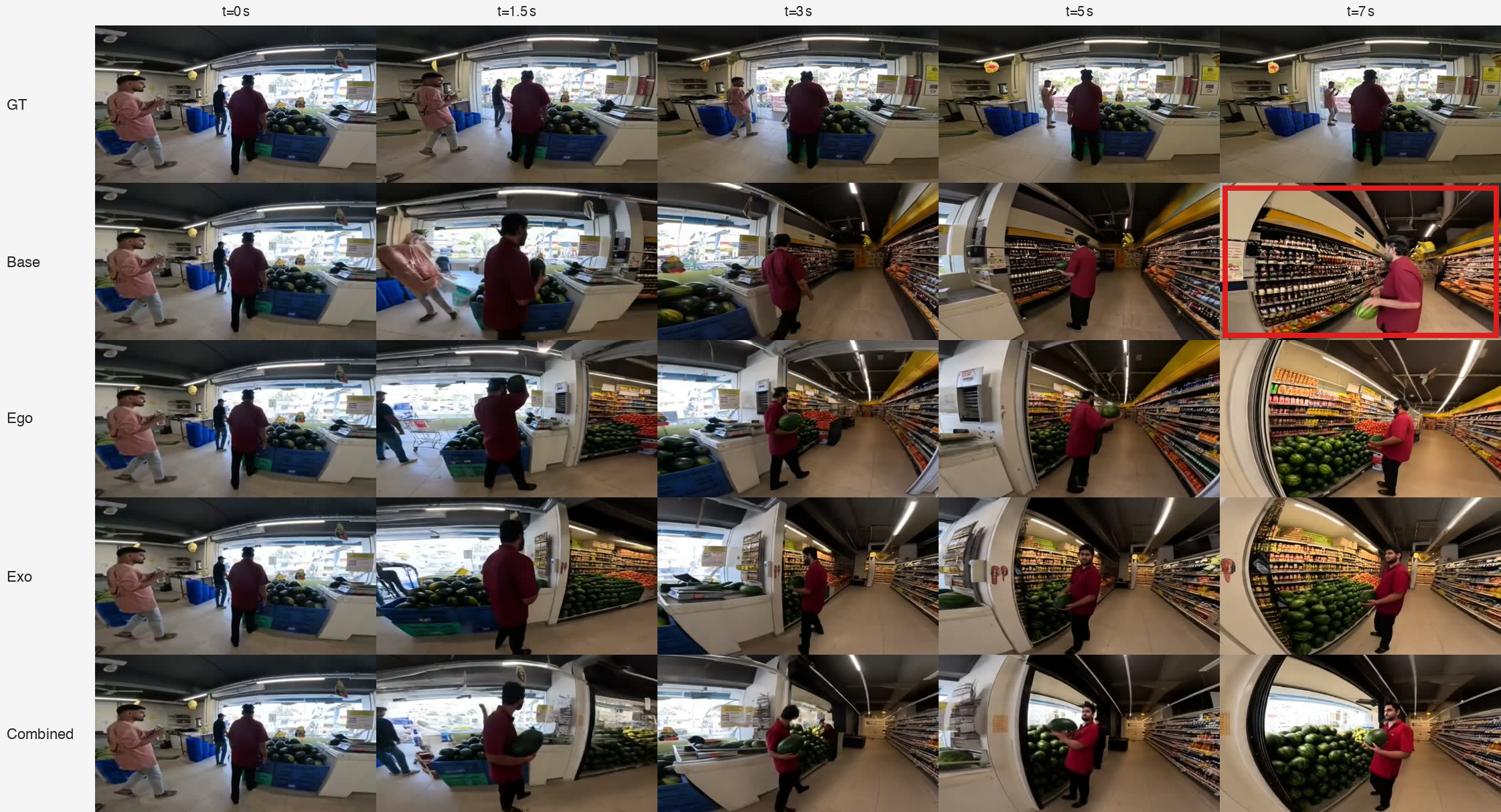}
\caption{\textbf{Hand-off watermelon.} A produce-area hand-off scene. Base drifts away from the multi-person produce display and ends at $t{=}7$~s with the person holding a watermelon in a single hand in the wrong section of the store. All three adapted configurations preserve the produce context and a two-handed grip consistent with the prompt.}
\label{fig:qual_handoff}
\end{figure*}

\nbf{Hand-off watermelon} \Cref{fig:qual_handoff} shows the conditioning frame of two people at a produce display, where the person is set up to receive a watermelon and hold it with both hands. The base configuration drifts away from the produce area within a few seconds and ends, at $t{=}7$~s, with a single person holding the watermelon awkwardly in one hand in what looks like an unrelated chocolate aisle. The egocentric-only, exocentric-only, and combined configurations all preserve the multi-person produce context and a two-handed grip on the watermelon consistent with the prompt. The contrast is consistent with the metric-level finding that base lacks the in-distribution retail priors that the adapted configurations acquire.

\begin{figure*}[t]
\centering
\includegraphics[width=\linewidth]{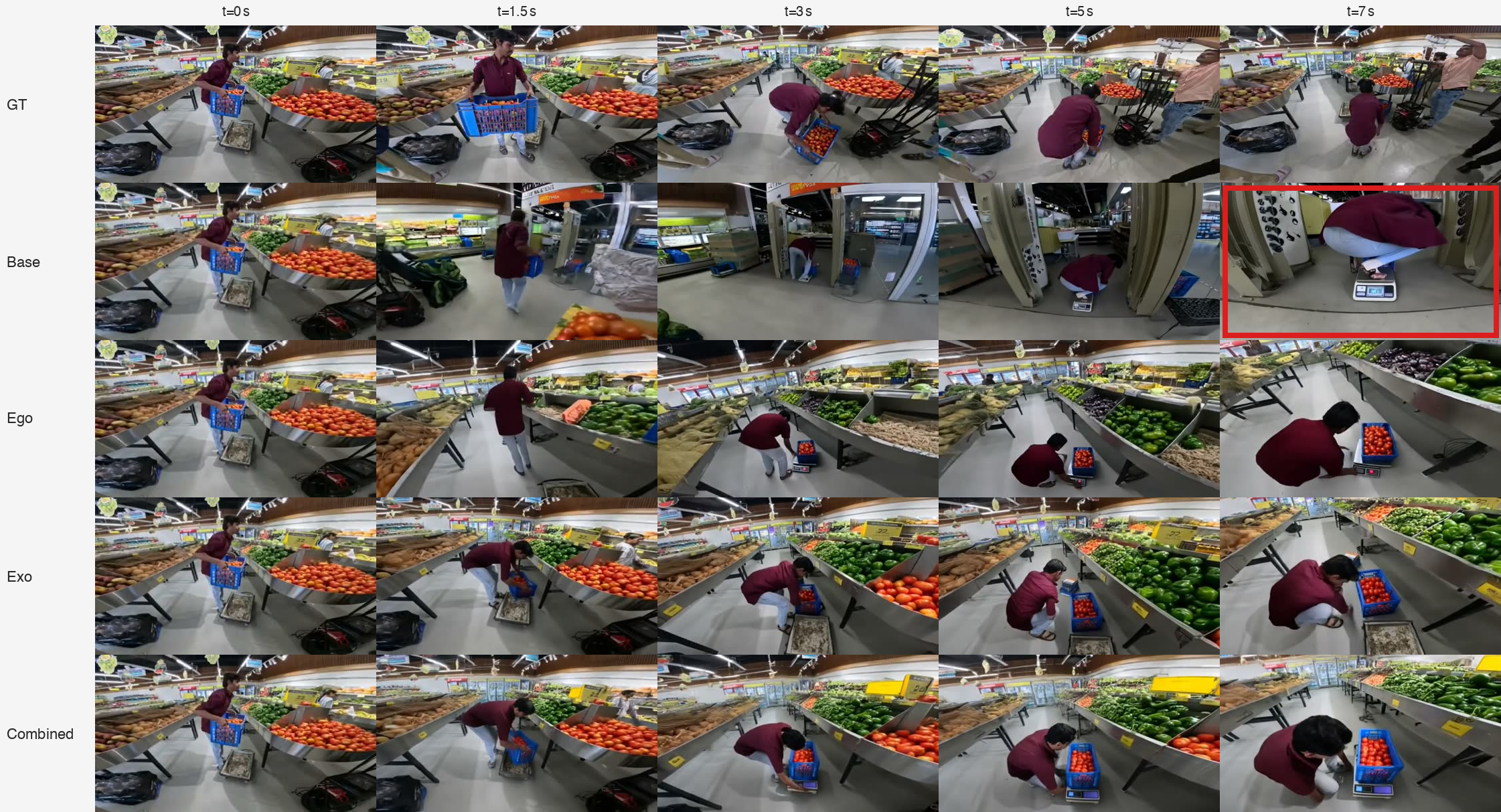}
\caption{\textbf{Weighing a tomato crate.} The prompt directs the person to place a blue tomato crate onto a floor weighing scale. Base never executes the bend-and-place action and ends with the person's body, rather than the crate, positioned on the scale. All three adapted configurations cleanly perform the squat-to-scale motion with the crate as the object that lands on the scale.}
\label{fig:qual_weigh}
\end{figure*}

\nbf{Weighing the tomato crate} \Cref{fig:qual_weigh} shows a weighing-station scene where the person is supposed to lift a blue crate of tomatoes and place it onto a floor scale. Base never produces the bend-and-place motion required by the prompt and ends with the person positioned on (or directly above) the scale rather than the crate. The egocentric-only and exocentric-only configurations both execute a recognizable squat-to-scale motion, and the combined configuration produces the most explicit final frame in which the crate is in contact with the scale. The base failure here is one of action grounding rather than scene quality, and is exactly the type of error that the per-pixel reconstruction metrics fail to penalize but that the semantic and distributional metrics in \cref{tab:com_vs_exo,tab:main} pick up.

\begin{figure*}[t]
\centering
\includegraphics[width=\linewidth]{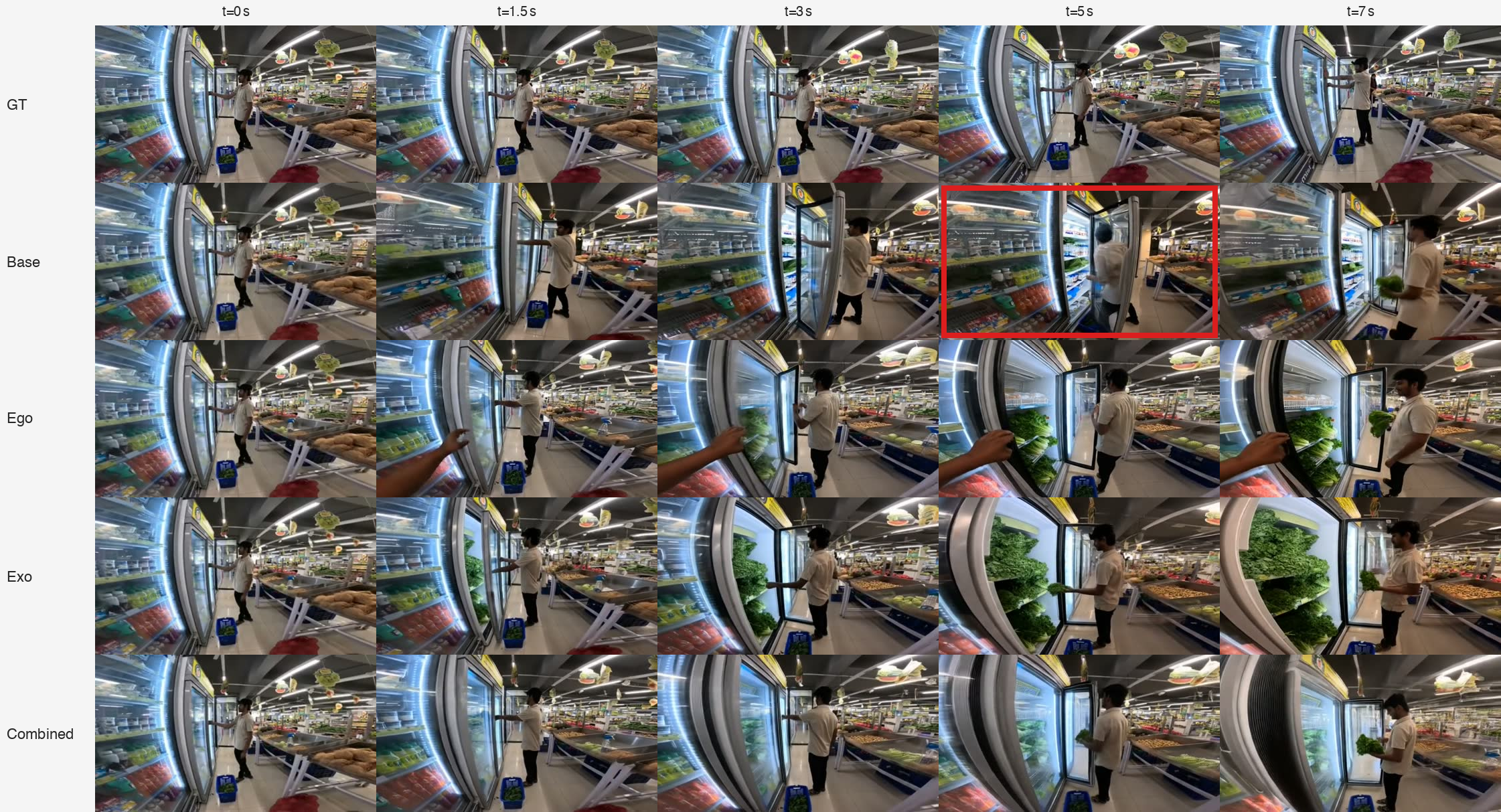}
\caption{\textbf{Opening a fridge to pick a vegetable.} Base produces a geometry-violating frame at $t{=}5$~s in which the fridge door appears to pass through the person's body. The three adapted configurations preserve the open-door geometry and a plausible reach for the vegetable inside the fridge.}
\label{fig:qual_fridge}
\end{figure*}

\nbf{Opening the fridge} \Cref{fig:qual_fridge} shows a refrigerator-side scene where the prompt asks the person to open the door and pick a vegetable from inside. Base produces an explicit geometry violation at $t{=}5$~s in which the fridge door appears to pass through the person's body. The adapted configurations preserve the open-door geometry and a plausible reach into the fridge. This is a low-level structural error in the base model that the adapted models avoid; it is the kind of mistake that disqualifies a world model for downstream embodied use, regardless of pixel-level similarity.

\begin{figure*}[t]
\centering
\includegraphics[width=\linewidth]{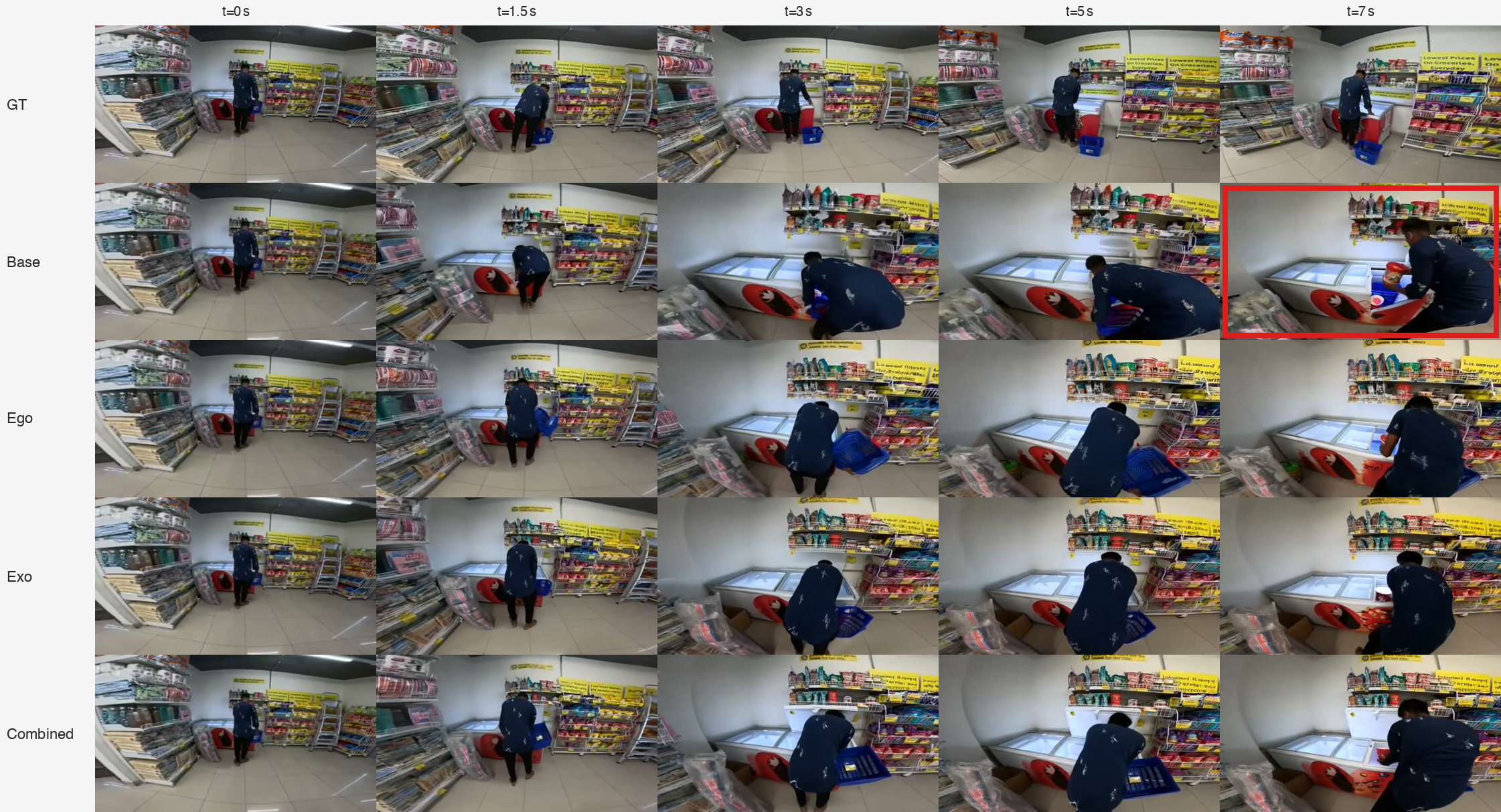}
\caption{\textbf{Opening a freezer to take an ice cream.} Base distorts the freezer structure as the rollout progresses; the adapted configurations preserve the freezer geometry, and exo-only produces the most physically sensible reach for the ice-cream tub.}
\label{fig:qual_freezer}
\end{figure*}

\nbf{Open the freezer, take ice cream} \Cref{fig:qual_freezer} shows a freezer-aisle scene where the person opens a freezer cabinet to take out an ice-cream tub. Base distorts the freezer structure as the rollout progresses, producing a final frame in which the freezer has lost its rectilinear geometry. The adapted configurations preserve the freezer cabinet structure throughout the rollout, and exo-only produces the most natural-looking reach into the freezer. This example also illustrates the headline finding numerically reported in \cref{tab:com_vs_exo}: exo-only and combined produce similar-quality outputs, but the small SSIM advantage of combined does not capture the additional semantic plausibility of exo-only's reach geometry.

\begin{figure*}[t]
\centering
\includegraphics[width=\linewidth]{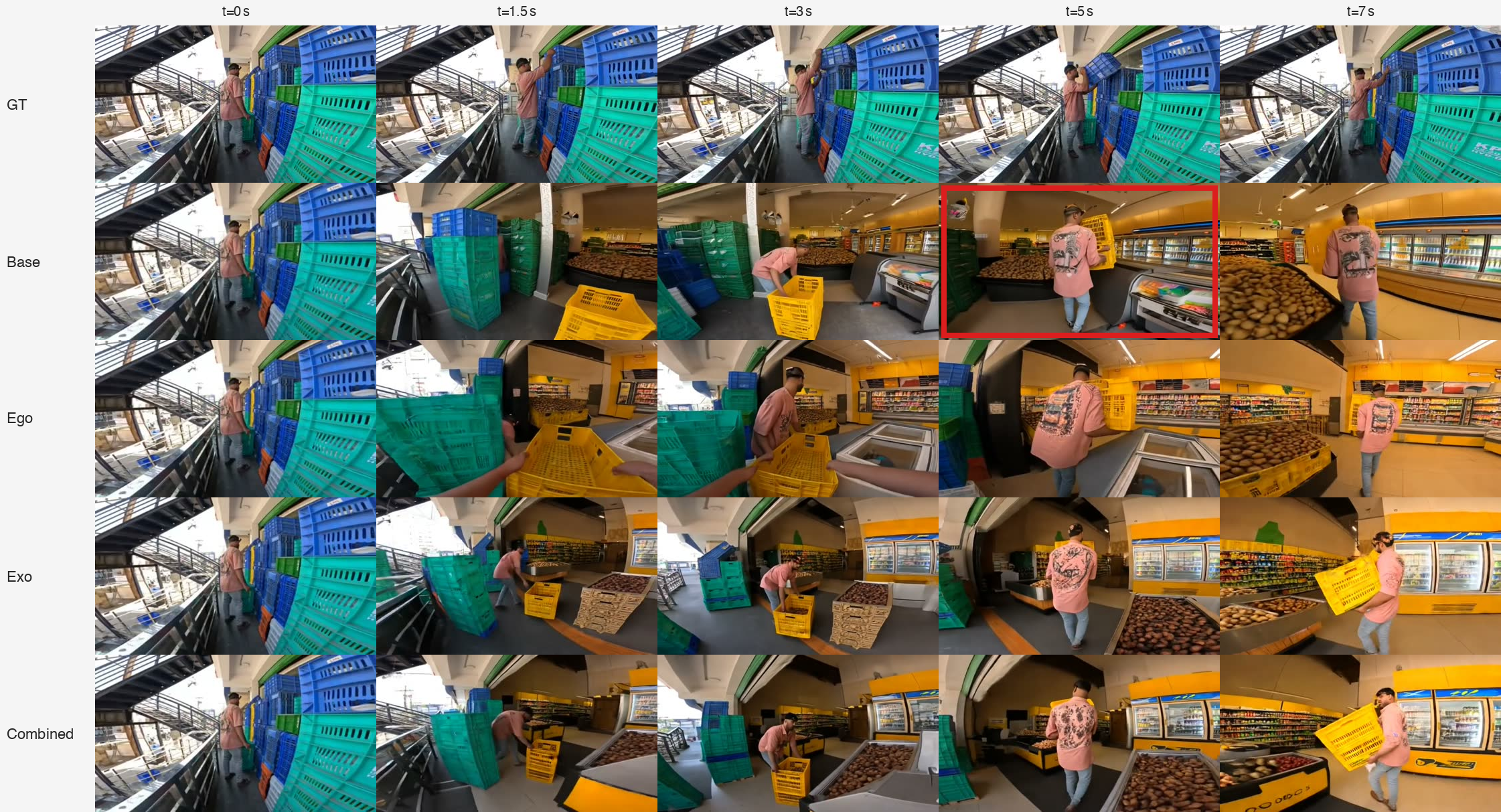}
\caption{\textbf{Carrying a crate through an aisle.} Base produces a hand--crate interpenetration at $t{=}5$~s: the person's hands appear to pass through the yellow crate while walking. The adapted configurations preserve a plausible two-hand carry posture.}
\label{fig:qual_carry}
\end{figure*}

\nbf{Carry the crate} \Cref{fig:qual_carry} shows the person walking down an aisle carrying a yellow crate. Base produces a hand--object interpenetration at $t{=}5$~s in which the person's hands appear to pass through the crate while walking. All three adapted configurations preserve a plausible two-hand carry posture in which the hands remain on the crate handles. As in the fridge example above, this is a hand--object physics failure that the per-pixel reconstruction metrics fail to penalize but that is unmistakable to a human observer.

\begin{figure*}[t]
\centering
\includegraphics[width=\linewidth]{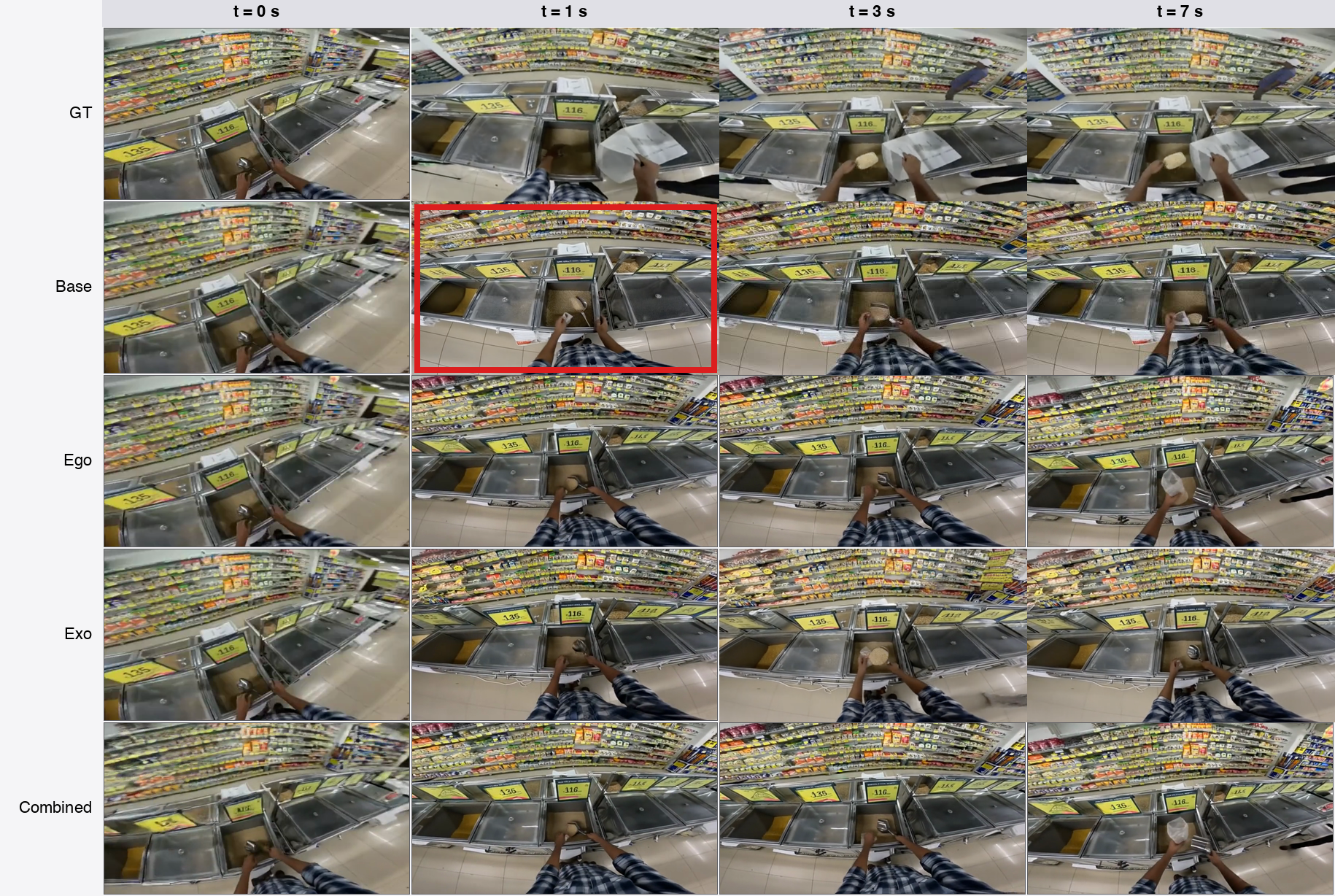}
\caption{\textbf{Picking up a metal scoop at the bulk-grain bins.} The conditioning frame shows the person standing at a row of open bulk-grain bins, ready to grasp a metal scoop. Base produces a frame at $t{=}1$~s in which the scoop drifts upward in mid-air with no hand around it (the agent--object grasp is lost); the three adapted configurations close the actor's hand on the scoop and lift it as a single rigid body.}
\label{fig:qual_scoop}
\end{figure*}

\nbf{Pick up the metal scoop} \Cref{fig:qual_scoop} shows the person standing at the bulk-grain bins, ready to take a metal scoop and use it to fill a plastic bag with grain. Base produces a hard agent--object grasp failure at $t{=}1$~s in which the scoop floats upward on its own with no hand around it. The egocentric-only, exocentric-only, and combined configurations all correctly close the person's hand around the scoop and lift it as a single rigid body. This is a hand--object physics failure in the earliest stage of the rollout that disqualifies the base output for downstream embodied use regardless of pixel-level similarity, and is exactly the type of error that the per-pixel reconstruction metrics under-penalize.

\begin{figure*}[t]
\centering
\includegraphics[width=\linewidth]{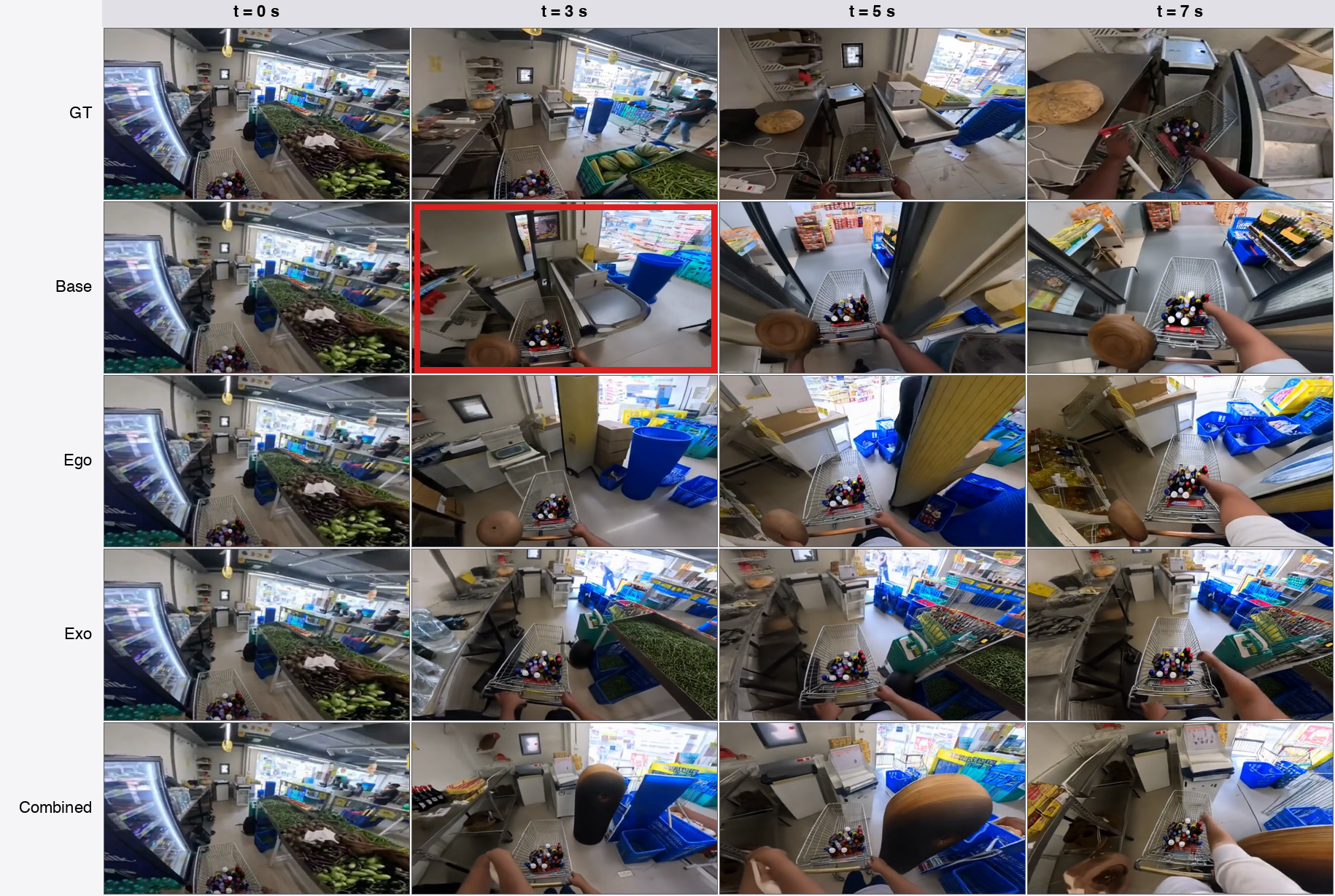}
\caption{\textbf{Pushing a cart through the aisle.} The conditioning frame shows the person pushing a cart with bottles down an aisle whose back wall is closed. Base produces a frame at $t{=}3$~s in which a spurious opening appears in the back wall of the aisle, exposing an unrelated store section that is not present in the conditioning frame. The three adapted configurations preserve the original aisle wall and product layout for the full rollout, with exocentric-only producing the most stable scene geometry.}
\label{fig:qual_cart}
\end{figure*}

\nbf{Push the cart down the aisle} \Cref{fig:qual_cart} shows a cart-navigation scene in which the person pushes a shopping cart loaded with small bottles down a backend aisle and then reaches into the cart with the right hand. The base configuration inserts a spurious opening in the back wall of the aisle at $t{=}3$~s, exposing an unrelated section of the store that was not visible in the conditioning frame; this is a scene-structure insertion failure rather than a pure pixel-level error. The three adapted configurations all preserve the original aisle geometry and the product layout on the side shelves throughout the rollout. The exocentric-only configuration produces the most stable scene up to $t{=}7$~s, consistent with the headline finding that exocentric-only adaptation specialises to wide-field scene structure.

\begin{figure*}[t]
\centering
\includegraphics[width=\linewidth]{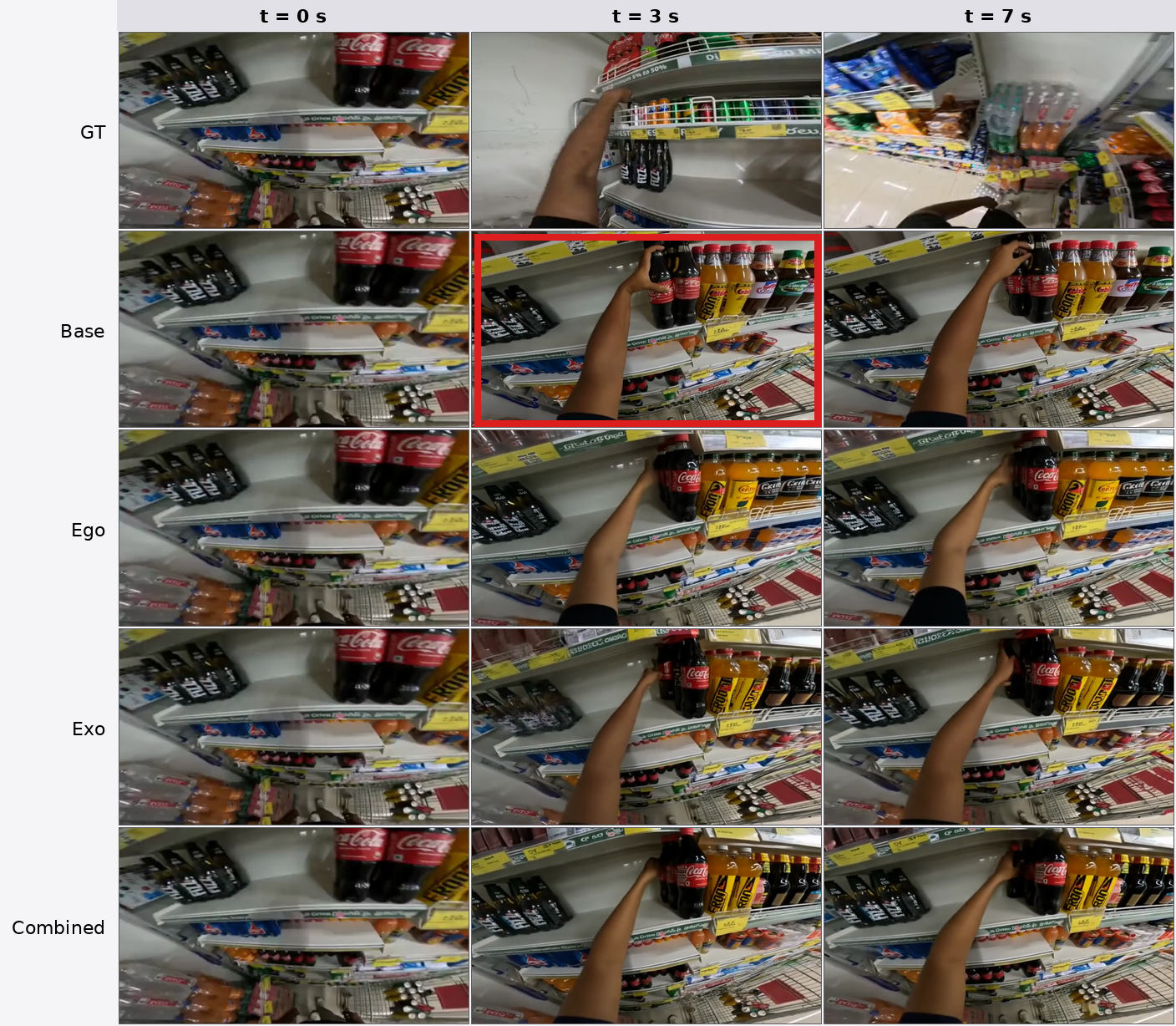}
\caption{\textbf{Reaching for a back-row bottle on the shelf.} The conditioning frame shows the person reaching for a black bottle at the back of a beverage shelf. Base produces an implausible grip pose around the bottle at $t{=}3$~s with no clear contact configuration; the three adapted configurations produce a recognisable reach-and-grasp consistent with picking up a back-row bottle.}
\label{fig:qual_bottle}
\end{figure*}

\nbf{Reach for the back-row bottle} \Cref{fig:qual_bottle} shows a beverage-cooler scene where the person reaches for a black bottle at the back of the shelf, picks it up, and places it in the front row. Base produces a hand-pose failure at $t{=}3$~s in which the hand wraps the bottle in an unnatural configuration with no clear contact between fingers and bottle body. The egocentric-only, exocentric-only, and combined configurations all produce a recognisable back-row reach-and-grasp consistent with the prompt. This is a fine-grained hand-pose failure that is invisible to scene-level distributional metrics but immediately apparent to a human observer.

\begin{figure*}[t]
\centering
\includegraphics[width=\linewidth]{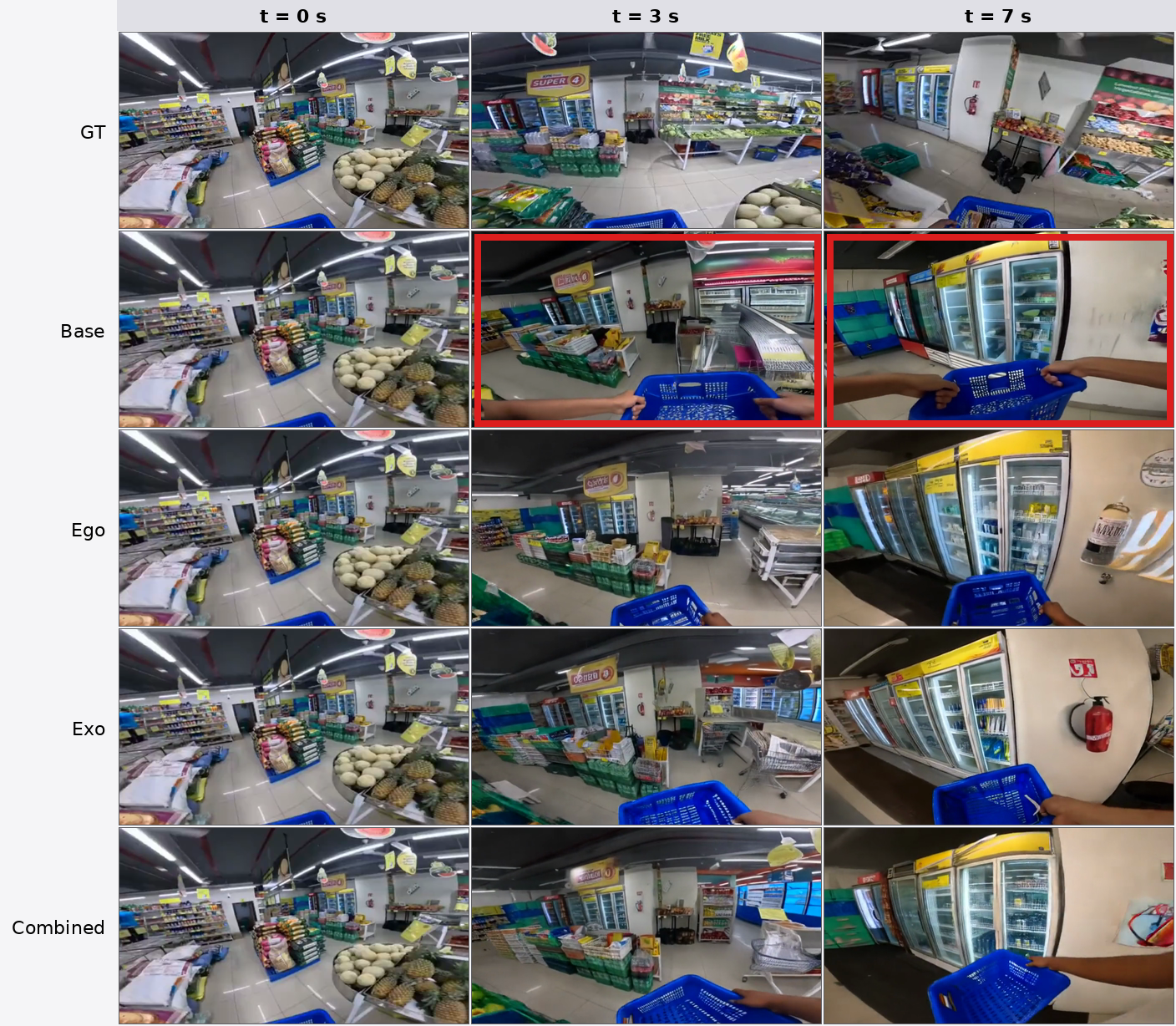}
\caption{\textbf{Carrying a basket through the aisles to the dairy section.} The conditioning frame shows the person walking with a blue shopping basket held with both hands. Base produces frames at $t{=}3$~s and $t{=}7$~s in which the person's hands enter the interior volume of the basket rather than gripping the handles; the three adapted configurations maintain a plausible two-hand basket carry posture throughout the rollout.}
\label{fig:qual_basket}
\end{figure*}

\nbf{Carry the basket to the dairy section} \Cref{fig:qual_basket} shows the person walking through the store aisles carrying a blue shopping basket with both hands, and approaching the dairy-section refrigerators. Base produces a hand--object interpenetration failure at $t{=}3$~s and again at $t{=}7$~s: the person's fingers appear to enter the inside volume of the basket rather than wrapping around the handles. The egocentric-only, exocentric-only, and combined configurations all preserve a plausible two-hand basket carry posture throughout the rollout. This is the same class of physics failure as the crate example above, replicated on a different agent--object pair, and it is consistently absent from all three adapted configurations.

\begin{figure*}[t]
\centering
\includegraphics[width=\linewidth]{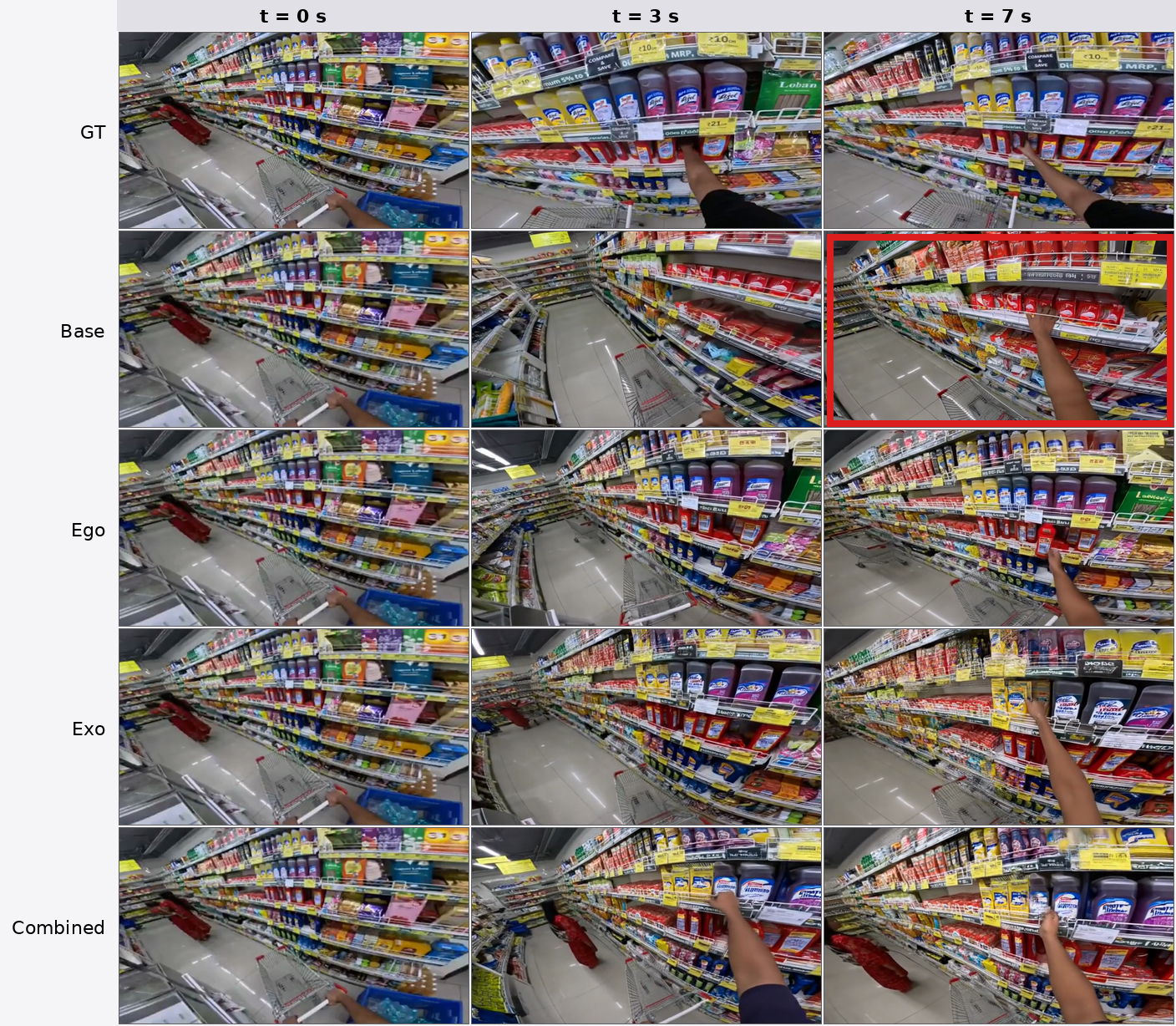}
\caption{\textbf{Inspecting a bottle of cleaning supplies.} The conditioning frame shows the person pushing a cart down a cleaning-supplies aisle. Base produces a frame at $t{=}7$~s in which the reaching hand penetrates the shelf guard rail; the three adapted configurations preserve a correct hand--shelf contact and a recognisable inspection pose.}
\label{fig:qual_shelf}
\end{figure*}

\nbf{Inspect the cleaning-supplies bottle} \Cref{fig:qual_shelf} shows an inspection scene in which the person pushes a cart down the aisle, stops, and reaches with the right hand to inspect a red bottle of cleaning supplies on the shelf. Base produces a contact-physics artifact at $t{=}7$~s in which the reaching hand penetrates the shelf guard rail rather than respecting it as a solid surface. The egocentric-only, exocentric-only, and combined configurations all preserve correct hand--shelf contact and produce a recognisable inspection pose. Together with the crate, basket, and fridge examples, this completes a family of four hand--surface and hand--object contact-physics failures in the base model, all of which the adapted configurations consistently avoid.

\subsection{Temporal Structure of the Adaptation Gap}
\label{sec:res_temporal}

\Cref{fig:lpips_vs_time} plots LPIPS as a function of rollout time on the test set. The absolute gap between the LoRA-adapted models and the pretrained base is widest at the shortest rollout time we sample ($t{=}0.5$ s, base mean LPIPS $0.542$ vs.\ exocentric-only $0.460$, a gap of $0.082$) and narrows steadily as $t$ grows, since all configurations drift away from the single ground-truth trajectory in the same way. \Cref{tab:near_horizon} reports the paired LPIPS gap at fixed time stamps within this near-horizon regime.

\begin{table}[t]
\centering
\small
\caption{Near-horizon LPIPS analysis on the held-out 200-clip test set. The paired LPIPS gap to the pretrained base widens from the unrestricted full-clip average ($+0.058$ for exocentric-only, $+0.049$ for combined, $+0.041$ for egocentric-only) to higher values in the near-horizon $t{\in}[1.0,2.0]$\,second window; \cref{fig:lpips_vs_time} in the appendix shows that the gap is largest at the shortest rollout time we sample ($t{=}0.5$\,s) and narrows monotonically thereafter.}
\label{tab:near_horizon}
\begin{tabular}{lcccc}
\toprule
\textbf{Time window} & \textbf{Configuration} & \textbf{LPIPS $\Delta$} & \textbf{Win} & \textbf{$p$} \\
\midrule
Unrestricted average & Egocentric-only & $+0.041$ & $90.5\%$ & ${<}10^{-27}$ \\
Unrestricted average & Exocentric-only & $\mathbf{+0.058}$ & $\mathbf{100\%}$ & ${<}\mathbf{10^{-34}}$ \\
Unrestricted average & Combined        & $+0.049$ & $94.5\%$ & ${<}10^{-31}$ \\
\midrule
$t{=}1.0$ second & Egocentric-only & $+0.061$ & $82.0\%$ & ${<}10^{-20}$ \\
$t{=}1.0$ second & Exocentric-only & $\mathbf{+0.075}$ & $\mathbf{89.0\%}$ & ${<}\mathbf{10^{-28}}$ \\
$t{=}1.0$ second & Combined        & $+0.068$ & $83.5\%$ & ${<}10^{-24}$ \\
\midrule
$t{=}2.0$ seconds & Egocentric-only & $+0.056$ & $87.5\%$ & ${<}10^{-25}$ \\
$t{=}2.0$ seconds & Exocentric-only & $\mathbf{+0.070}$ & $\mathbf{96.0\%}$ & ${<}\mathbf{10^{-32}}$ \\
$t{=}2.0$ seconds & Combined        & $+0.064$ & $88.5\%$ & ${<}10^{-27}$ \\
\bottomrule
\end{tabular}
\end{table}

The near-horizon LPIPS gap to the pretrained base is largest at the shortest rollout time we sample ($t{=}0.5$\,s, see \cref{fig:lpips_vs_time} in the appendix) and remains substantial throughout the $t{\in}[1.0,2.0]$\,s window. At $t{=}1.0$\,s the exocentric-only configuration reaches a paired LPIPS gap of $+0.075$ at $89.0\%$ paired win-rate ($p{<}10^{-28}$) and the combined configuration $+0.068$ at $83.5\%$ ($p{<}10^{-24}$). At $t{=}2.0$\,s the gap remains substantial ($+0.064$ to $+0.070$ at $87.5{-}96.0\%$ paired win-rate). At longer rollout horizons all configurations drift away from the single ground-truth trajectory in the same way, and the absolute LPIPS gap narrows; we omit those time stamps from the table for clarity and report the full curve in the appendix.

The near-horizon window has a clear interpretive meaning: it is the regime in which the world model's prediction quality matters most for embodied use, since downstream planning and policy verification typically rely on rollouts of one to three seconds~\cite{gaia2}. Our results indicate that LoRA adaptation is most beneficial precisely in this regime.

\subsection{Ablations and Sanity Checks}
\label{sec:ablations}

\nbf{Time-aligned vs.\ frame-index-based sampling} A naive LPIPS implementation would compare generated and ground-truth frames by index, mis-aligning the generated 24~FPS output against ground-truth at 4~FPS. Switching to the wall-clock time-aligned protocol of \cref{sec:metrics} increases the paired LPIPS gap on our test set by approximately $3.9\times$ over the index-based comparison. We recommend the time-aligned protocol for any video evaluation in which the generated and reference frame rates differ.

\nbf{Adapter merge verification} Across all 36 transformer blocks the post-merge weight differs from the pretrained weight in $98.83\%$ of its entries, with per-tensor maximum delta magnitude of $17.8\%$ of the base tensor magnitude. This excludes a class of failure modes in which the adapter loads silently as a zero increment due to a Diffusers PEFT name mismatch.

\nbf{Seed variance} The seed-induced spread on the combined configuration (\cref{tab:val_loss}) is at least an order of magnitude smaller than the gap between any LoRA configuration and the base on every paired metric, supporting the interpretation that observed configuration-to-configuration ordering is not driven by optimization noise.

\nbf{Modern metric sensitivity} On our test set, the DreamSim paired improvement for the exocentric-only configuration corresponds to a relative reduction of $23.6\%$ from the base mean, approximately $2.7{\times}$ the LPIPS relative reduction of $8.7\%$. On the distributional side, JEDi shows a $33.5\%$ relative reduction for exocentric-only versus $17.7\%$ for R3D-Fr\'echet, a $1.9\times$ amplification. In both pairings the modern metric agrees with the classical metric on the configuration ranking but provides a more sensitive measurement of the adaptation effect, consistent with the design of DreamSim~\cite{fu2023dreamsim} and JEDi~\cite{luo2024beyondfvd} to capture aspects of similarity that the classical metrics miss.

\section{Conclusion}
\label{sec:conclusion}

We introduced \textbf{RetailSMV} (\emph{Retail Synchronized Multi-View}), a retail video corpus of $32{,}105$ captioned clips with synchronized egocentric and exocentric capture of the same store-staff activities across five real-world supermarkets, and used it to conduct a controlled study of view-stratified Low-Rank Adaptation for foundation video world models. By training three matched LoRA configurations of Cosmos3-Nano on RetailSMV under identical hyperparameters and optimization budget, and by evaluating them on a held-out test set with a seven-metric suite under a strict paired statistical protocol, we have arrived at several conclusions of practical interest for the world-model adaptation community.

\nbf{Adaptation is statistically uniform} The reduction in validation diffusion loss is large ($\sim$$2.8\times$), uniform across configurations, and achieves perfect paired ordering ($p\ll 0.001$). On this aggregate signal alone there is no ambiguity that the pretrained foundation video model benefits substantially from domain-specific LoRA adaptation. Validation loss alone, however, does not discriminate between view-stratified data choices; the configuration differences emerge when generation quality is measured directly.

\nbf{Exocentric-only adaptation matches or exceeds combined adaptation, with significant gains on LPIPS, PSNR, and DreamSim} Our headline finding is that exocentric-only training, using only the $15{,}985$ exocentric clips of the synchronized ego$+$exo corpus, matches or exceeds the combined configuration on six of the seven point estimates we report. The exocentric-only adapter wins on R3D-Fréchet ($35.14$ vs.\ combined $36.20$ vs.\ base $42.69$), on JEDi ($0.829$ vs.\ combined $0.892$ vs.\ base $1.246$, a $33.5\%$ relative reduction), on LPIPS (a paired reduction of $0.058$ at $100\%$ paired win-rate vs.\ combined's $0.049$ at $94.5\%$), on PSNR (an improvement of $+0.575$\,dB vs.\ $+0.500$\,dB), on Hessel CLIPScore ($+0.018$ vs.\ $+0.017$), and on DreamSim (a reduction of $0.062$ vs.\ $0.046$), even though the combined configuration also sees the $16{,}120$ egocentric clips. The combined configuration edges exo only on SSIM, by $0.0012$. Direct paired tests between the two adapted configurations (\cref{tab:com_vs_exo}) confirm that combined is significantly \emph{worse} than exo on LPIPS, PSNR, and DreamSim, and the two distributional metrics (R3D-Fréchet and JEDi) agree on the ranking under either feature backbone. The corresponding combined-vs-egocentric tests (\cref{tab:com_vs_ego}) show the asymmetric counterpart: combined is significantly \emph{better} than ego-only on LPIPS ($p{=}0.002$), PSNR ($p{<}10^{-6}$), and SSIM ($p{<}10^{-14}$). Adding the exocentric subset to the egocentric subset therefore helps, whereas adding the egocentric subset to the exocentric subset hurts. When synchronized ego/exo capture is available, single-view exocentric training is therefore a strong default; combined ego$+$exo training does not improve over exo-only training on this corpus. We hypothesize that this reflects the wider field of view and stable reference frame of the exocentric camera, which align well with the inductive prior of a diffusion video model that was pretrained primarily on stable third-person footage. What matters is in-distribution coverage of the deployment viewpoint, not raw clip count.

\nbf{Egocentric data transfers, but ranks third} The egocentric-only adapter trained on $16{,}120$ egocentric clips improves over the pretrained base on every metric at $p{<}10^{-8}$, including LPIPS ($+0.041$ at $90.5\%$ paired win-rate), CLIPScore ($+0.017$ at $72.5\%$), and DreamSim ($+0.043$ at $78.0\%$). The transfer is substantive but consistently weaker than the in-distribution exocentric-only adapter. This pattern may inform data-collection priorities for future embodied world models in retail-like deployment domains.

\nbf{The adaptation gap is concentrated in the near-horizon prediction window} The absolute LPIPS gap to the pretrained base is largest at the shortest rollout time we sample ($t{=}0.5$ s, base mean LPIPS $0.542$ vs.\ exocentric-only $0.460$, a gap of $0.082$) and narrows steadily as rollout time grows, since all configurations drift away from the single ground-truth trajectory in the same way. In the $t{\in}[1.0,2.0]$\,s window the paired gap is still substantial: at $t{=}1.0$\,s the exocentric-only configuration reaches a paired LPIPS improvement of $+0.075$ at $89\%$ paired win-rate, approximately $1.3\times$ the full-clip average; the combined and egocentric-only configurations exhibit the same near-horizon amplification at $+0.068$ and $+0.061$ respectively. The near-horizon window is precisely the regime in which world models are most directly used for embodied control~\cite{gaia2}. We recommend reporting the LPIPS gap at fixed near-horizon time stamps in addition to the full-clip average as a standard reporting convention.

\nbf{Modern metrics are more sensitive} The DreamSim paired improvement for the exocentric-only configuration is approximately $2.7\times$ the LPIPS paired improvement in relative terms ($23.6\%$ vs.\ $8.7\%$), while the two metrics produce the same configuration ranking; on the distributional side, JEDi's $33.5\%$ relative reduction is roughly $1.9\times$ the corresponding R3D-Fr\'echet reduction ($17.7\%$), again with identical configuration ordering. This is consistent with the design of DreamSim and JEDi to capture mid-level semantic and feature-distribution similarity that the classical metrics miss~\cite{fu2023dreamsim,luo2024beyondfvd}. We report JEDi as a recent distributional video-quality metric \emph{complementary} to R3D-Fréchet; the conclusions in this paper do not rely on JEDi alone and are supported by R3D-Fréchet and DreamSim as well. We recommend that domain-specific video evaluations report DreamSim alongside (or in place of) LPIPS, and a V-JEPA-based MMD alongside an I3D/R3D Fréchet distance.

\subsection*{Limitations}

Our study has four limitations we believe deserve explicit mention.

First, we adapt one base model family (Cosmos3-Nano). A natural extension is to evaluate whether the qualitative findings transfer to other foundation video models such as Cosmos-Predict~2.5, Sora, and Stable Video Diffusion. A parallel adaptation effort to Cosmos-Predict~2.5 is ongoing.

Second, our Fréchet-style metric uses a ResNet-3D Kinetics backbone (R3D-Fréchet) rather than the canonical I3D backbone, so absolute values are not directly comparable to I3D-FVD values in the broader literature; see \cref{sec:appendix_notes} for details and a recipe for recomputing with the canonical I3D backbone.

Third, we do not conduct a human evaluation. The metrics community has documented that human judgment remains the gold standard for video generation quality~\cite{fu2023dreamsim,luo2024beyondfvd}. Our paired statistical protocol provides a transparent basis for any subsequent human-versus-automatic alignment study, and the modern metric we adopt (DreamSim) is designed precisely to narrow the gap to human judgment, but a direct human comparison would strengthen the conclusions.

Fourth, we use image-and-text conditioning. The natural use case for a world model in embodied control is action conditioning, where the model rolls out a counterfactual based on a proposed action sequence. Cosmos3-Nano does not natively support action conditioning; adding it is a promising direction for downstream evaluation as a true simulator for embodied control.

Beyond these specific limitations, the broader takeaway is that view-stratified studies, paired statistics, and modern perceptual metrics together support reproducible, well-controlled studies of parameter-efficient video world model adaptation. We hope the protocol and findings presented here are useful starting points for the next generation of domain-specialized generative simulators.

\clearpage
\bibliographystyle{plainnat}
\bibliography{11_references}

@String(CVPR  = {CVPR})

@String(ICCV  = {ICCV})

@String(ECCV  = {ECCV})

@String(NIPS  = {NeurIPS})

@String(BMVC  =	{BMVC})

@String(TIP   = {IEEE TIP})

@String(ICLR  = {ICLR})

@article{cosmos_reason2,
  title={Cosmos-Reason1 and Cosmos-Reason2: Reasoning Foundation Models for Physical Common Sense},
  author={NVIDIA},
  journal={arXiv preprint},
  year={2025}
}

@article{cosmos_reason1,
  title={Cosmos-Reason1: From Physical Common Sense To Embodied Reasoning},
  author={NVIDIA},
  journal={arXiv preprint},
  year={2025}
}

@inproceedings{adamw,
      title={Decoupled Weight Decay Regularization},
      author={Loshchilov, Ilya and Hutter, Frank},
      booktitle=ICLR,
      year={2019},
      eprint={1711.05101},
      archivePrefix={arXiv},
      primaryClass={cs.LG},
      url={https://arxiv.org/abs/1711.05101}, 
}

@article{groot_n1,
  title={{GR00T} N1: An Open Foundation Model for Generalist Humanoid Robots},
  author={NVIDIA},
  journal={arXiv preprint},
  year={2025}
}

@inproceedings{rt2,
  title={{RT-2}: Vision-Language-Action Models Transfer Web Knowledge to Robotic Control},
  author={Brohan, Anthony and others},
  booktitle={Conference on Robot Learning (CoRL)},
  year={2023}
}

@article{palme,
  title={{PaLM-E}: An Embodied Multimodal Language Model},
  author={Driess, Danny and others},
  journal={arXiv preprint arXiv:2303.03378},
  year={2023}
}

@article{pi0,
  title={$\pi_0$: A Vision-Language-Action Flow Model for General Robot Control},
  author={Black, Kevin and others},
  journal={arXiv preprint arXiv:2410.24164},
  year={2024}
}

@article{robovqa,
  title={{RoboVQA}: Multimodal Long-Horizon Reasoning for Robotics},
  author={Sermanet, Pierre and Ding, Tianli and Zhao, Jeffrey and Xia, Fei and
          Dwibedi, Debidatta and Gopalakrishnan, Keerthana and Chan, Christine and
          Dulac-Arnold, Gabriel and Maddineni, Sharath and Jain, Nikhil and
          Xu, Peng and Yuan, Yunfei and others},
  journal={arXiv preprint arXiv:2311.00899},
  year={2023}
}

@inproceedings{ego4d,
  title={{Ego4D}: Around the World in 3,000 Hours of Egocentric Video},
  author={Grauman, Kristen and others},
  booktitle=CVPR,
  year={2022}
}

@article{egoexo4d,
  title={{Ego-Exo4D}: Understanding Skilled Human Activity from First- and Third-Person Perspectives},
  author={Grauman, Kristen and others},
  journal={arXiv preprint arXiv:2311.18259},
  year={2024}
}

@inproceedings{epic_kitchens,
  title={Scaling Egocentric Vision: The {EPIC-KITCHENS} Dataset},
  author={Damen, Dima and others},
  booktitle=ECCV,
  year={2018}
}

@inproceedings{lora,
  title={{LoRA}: Low-Rank Adaptation of Large Language Models},
  author={Hu, Edward J and others},
  booktitle=ICLR,
  year={2022}
}

@article{qlora,
  title={{QLoRA}: Efficient Finetuning of Quantized Language Models},
  author={Dettmers, Tim and others},
  journal={arXiv preprint arXiv:2305.14314},
  year={2023}
}

@inproceedings{arrow_of_time,
  title={Arrow of Time and its Reversal on the {IBM} Quantum Computer},
  author={Wei, Dongling and others},
  booktitle={Scientific Reports},
  year={2019},
  note={Original concept: Pickup et al., Arrow of Time in Videos, BMVC 2014}
}

@inproceedings{jigsaw,
  title={Unsupervised Visual Representation Learning by Context Prediction},
  author={Doersch, Carl and Gupta, Abhinav and Efros, Alexei A},
  booktitle=ICCV,
  year={2015}
}

@inproceedings{merl_shopping,
  title={A Multi-Stream Bi-Directional Recurrent Neural Network for Fine-Grained Action Detection},
  author={Singh, Bharat and Marks, Tim K and Jones, Michael and Tuzel, Oncel and Shao, Ming},
  booktitle=CVPR,
  year={2016}
}

@misc{retailvision,
  title={{RetailVision} Workshop Series},
  author={{RetailVision Organizers}},
  howpublished={\url{https://retailvisionworkshop.github.io}},
  year={2020--2025},
  note={Annual workshop at CVPR/ICCV, 2020--2025}
}

@misc{gemini_robotics,
  author = {Gemini-Robotics-ER-1.5},
  title = {Google Gemini ER 1.5},
  url = {https://ai.google.dev/gemini-api/docs/robotics-overview},
  year = {2025}
}

@inproceedings{saribench,
  title={{Sari Sandbox}: A Virtual Retail Store Environment for Embodied {AI} Agents},
  author={Gajo, Justine and Merales, Aldrin and Escarcha, Timothy and Molina, Marco and Nartea, Antonio and Maminta, Andrei and Roldan, Joshua and Atienza, Rowel},
  booktitle={Proceedings of the IEEE/CVF International Conference on Computer Vision Workshops (ICCVW), RetailVision},
  year={2025},
  note={arXiv:2508.00400}
}

@inproceedings{retailaction,
  title={{RetailAction}: Dataset for Multi-View Spatio-Temporal Localization of Human--Object Interactions in Retail Environments},
  author={Mazzini, Davide and others},
  booktitle={Proceedings of the IEEE/CVF International Conference on Computer Vision Workshops (ICCVW), RetailVision},
  year={2025}
}

@article{agibot,
  title={AgiBot World Colosseo: A Large-scale Manipulation Platform for Scalable and Intelligent Embodied Systems},
  author={AgiBot World Contributors},
  journal={arXiv preprint arXiv:2503.06669},
  year={2025}
}

@article{cosmos_paper,
  title={{Cosmos World Foundation Model Platform for Physical AI}},
  author={{NVIDIA}},
  journal={arXiv preprint arXiv:2501.03575},
  year={2025}
}

@misc{sora2024,
  title={Video generation models as world simulators},
  author={{OpenAI}},
  year={2024},
  howpublished={openai.com/sora}
}

@article{svd2023,
  title={Stable Video Diffusion: Scaling Latent Video Diffusion Models to Large Datasets},
  author={Blattmann, Andreas and Dockhorn, Tim and Kulal, Sumith and others},
  journal={arXiv preprint arXiv:2311.15127},
  year={2023}
}

@misc{opensora2024,
  title={Open-Sora: Democratizing Efficient Video Production for All},
  author={Zheng, Zangwei and others},
  year={2024},
  howpublished={github.com/hpcaitech/Open-Sora}
}

@article{moviegen2024,
  title={Movie Gen: A Cast of Media Foundation Models},
  author={{Meta GenAI}},
  journal={arXiv preprint arXiv:2410.13720},
  year={2024}
}

@article{gaia1,
  title={{GAIA-1}: A Generative World Model for Autonomous Driving},
  author={Hu, Anthony and Russell, Lloyd and Yeo, Hudson and others},
  journal={arXiv preprint arXiv:2309.17080},
  year={2023}
}

@article{gaia2,
  title={{GAIA-2}: A Controllable Multi-View Generative World Model for Autonomous Driving},
  author={{Wayve}},
  journal={arXiv preprint arXiv:2503.20523},
  year={2025}
}

@inproceedings{drivedreamer,
  title={{DriveDreamer}: Towards Real-world-driven World Models for Autonomous Driving},
  author={Wang, Xiaofeng and others},
  booktitle=ECCV,
  year={2024}
}

@article{vista2024,
  title={{Vista}: A Generalizable Driving World Model with High Fidelity and Versatile Controllability},
  author={Gao, Shenyuan and others},
  journal={arXiv preprint arXiv:2405.17398},
  year={2024}
}

@article{ha2018worldmodels,
  title={World Models},
  author={Ha, David and Schmidhuber, J{\"u}rgen},
  journal={arXiv preprint arXiv:1803.10122},
  year={2018}
}

@article{gao2025worldmodel,
  title={Survey of Generative World Models for Embodied AI},
  author={Gao, Shenyuan and Yang, Jiazhi and Chen, Li},
  journal={arXiv preprint arXiv:2502.00060},
  year={2025}
}

@inproceedings{vbench,
  title={{VBench}: Comprehensive Benchmark Suite for Video Generative Models},
  author={Huang, Ziqi and He, Yinan and Yu, Jiashuo and others},
  booktitle=CVPR,
  year={2024}
}

@article{vbench2,
  title={{VBench-2.0}: Advancing Video Generation Benchmark Suite for Intrinsic Faithfulness},
  author={Huang, Ziqi and others},
  journal={arXiv preprint arXiv:2503.21755},
  year={2025}
}

@article{fvd_original,
  title={Towards Accurate Generative Models of Video: A New Metric \& Challenges},
  author={Unterthiner, Thomas and others},
  journal={arXiv preprint arXiv:1812.01717},
  year={2018}
}

@inproceedings{i3d,
  title={Quo Vadis, Action Recognition? A New Model and the {Kinetics} Dataset},
  author={Carreira, Jo{\~a}o and Zisserman, Andrew},
  booktitle=CVPR,
  year={2017}
}

@inproceedings{lpips,
  title={The Unreasonable Effectiveness of Deep Features as a Perceptual Metric},
  author={Zhang, Richard and Isola, Phillip and Efros, Alexei A and Shechtman, Eli and Wang, Oliver},
  booktitle=CVPR,
  year={2018}
}

@article{ssim,
  title={Image Quality Assessment: from Error Visibility to Structural Similarity},
  author={Wang, Zhou and Bovik, Alan C and Sheikh, Hamid R and Simoncelli, Eero P},
  journal=TIP,
  year={2004}
}

@article{clipscore,
  title={{CLIPScore}: A Reference-free Evaluation Metric for Image Captioning},
  author={Hessel, Jack and Holtzman, Ari and Forbes, Maxwell and others},
  journal={arXiv preprint arXiv:2104.08718},
  year={2021}
}

@inproceedings{fu2023dreamsim,
  title={{DreamSim}: Learning New Dimensions of Human Visual Similarity using Synthetic Data},
  author={Fu, Stephanie and Tamir, Netanel and Sundaram, Shobhita and others},
  booktitle=NIPS,
  year={2023}
}

@inproceedings{luo2024beyondfvd,
  title={Beyond {FVD}: Enhanced Evaluation Metrics for Video Generation Quality},
  author={Luo, Ge Ya and Favero, Gian Mario and Luo, Zhi-Hao and others},
  booktitle=NIPS,
  year={2024}
}

@inproceedings{rectifiedflow,
  title={Flow Straight and Fast: Learning to Generate and Transfer Data with Rectified Flow},
  author={Liu, Xingchao and Gong, Chengyue and Liu, Qiang},
  booktitle=ICLR,
  year={2023}
}

@inproceedings{aria,
  title={{Aria} Everyday Activities Dataset},
  author={{Meta Reality Labs}},
  booktitle=CVPR,
  year={2024}
}

@inproceedings{charades_ego,
  title={Actor and Observer: Joint Modeling of First and Third-person Videos},
  author={Sigurdsson, Gunnar A and Gupta, Abhinav and Schmid, Cordelia and others},
  booktitle=CVPR,
  year={2018}
}

@article{rouhi2026prism,
  title={{PRISM}: A Multi-View Multi-Capability Retail Video Dataset for Embodied Vision-Language Models},
  author={Rouhi, Amirreza and Sakurikar, Parikshit and Reddy, Satya Sai and Menga, Narsimha and Govil, Anirudh and Chittajallu, Sri Harsha and Aggarwal, Rajat and Namboodiri, Anoop and Reddi, Sashi},
  journal={arXiv preprint arXiv:2603.29281},
  year={2026}
}

@inproceedings{ddpm,
  title={Denoising Diffusion Probabilistic Models},
  author={Ho, Jonathan and Jain, Ajay and Abbeel, Pieter},
  booktitle=NIPS,
  year={2020}
}

@inproceedings{ddim,
  title={Denoising Diffusion Implicit Models},
  author={Song, Jiaming and Meng, Chenlin and Ermon, Stefano},
  booktitle=ICLR,
  year={2021}
}

@inproceedings{flowmatching,
  title={Flow Matching for Generative Modeling},
  author={Lipman, Yaron and Chen, Ricky T. Q. and Ben-Hamu, Heli and Nickel, Maximilian and Le, Matt},
  booktitle=ICLR,
  year={2023}
}

@inproceedings{unipc,
  title={{UniPC}: A Unified Predictor-Corrector Framework for Fast Sampling of Diffusion Models},
  author={Zhao, Wenliang and Bai, Lujia and Rao, Yongming and Zhou, Jie and Lu, Jiwen},
  booktitle=NIPS,
  year={2023}
}

@inproceedings{makeavideo,
  title={{Make-A-Video}: Text-to-Video Generation without Text-Video Data},
  author={Singer, Uriel and Polyak, Adam and Hayes, Thomas and others},
  booktitle=ICLR,
  year={2023}
}

@inproceedings{animatediff,
  title={{AnimateDiff}: Animate Your Personalized Text-to-Image Diffusion Models without Specific Tuning},
  author={Guo, Yuwei and Yang, Ceyuan and Rao, Anyi and others},
  booktitle=ICLR,
  year={2024}
}

@inproceedings{dynamicrafter,
  title={{DynamiCrafter}: Animating Open-domain Images with Video Diffusion Priors},
  author={Xing, Jinbo and Xia, Menghan and Zhang, Yong and others},
  booktitle=ECCV,
  year={2024}
}

@inproceedings{videocrafter2,
  title={{VideoCrafter2}: Overcoming Data Limitations for High-Quality Video Diffusion Models},
  author={Chen, Haoxin and Zhang, Yong and Cun, Xiaodong and others},
  booktitle=CVPR,
  year={2024}
}

@inproceedings{cogvideox,
  title={{CogVideoX}: Text-to-Video Diffusion Model with An Expert Transformer},
  author={Yang, Zhuoyi and Teng, Jiayan and Zheng, Wendi and others},
  booktitle=ICLR,
  year={2025}
}

@article{hunyuanvideo,
  title={{HunyuanVideo}: A Systematic Framework For Large Video Generative Models},
  author={Kong, Weijie and Tian, Qi and Zhang, Zijian and others},
  journal={arXiv:2412.03603},
  year={2024}
}

@inproceedings{lumiere,
  title={{Lumiere}: A Space-Time Diffusion Model for Video Generation},
  author={Bar-Tal, Omer and Chefer, Hila and Tov, Omer and Brooks, Tim and Hertz, Amir and Dekel, Tali and Mosseri, Inbar},
  booktitle={ACM SIGGRAPH Asia},
  year={2024}
}

@inproceedings{videopoet,
  title={{VideoPoet}: A Large Language Model for Zero-Shot Video Generation},
  author={Kondratyuk, Dan and Yu, Lijun and Gu, Xiuye and others},
  booktitle={ICML},
  year={2024}
}

@inproceedings{genie,
  title={{Genie}: Generative Interactive Environments},
  author={Bruce, Jake and Dennis, Michael D and Edwards, Ashley and Parker-Holder, Jack and Shi, Yuge and Hughes, Edward and others},
  booktitle={ICML},
  year={2024}
}

@inproceedings{unisim,
  title={Learning Interactive Real-World Simulators},
  author={Yang, Mengjiao and Du, Yilun and Ghasemipour, Kamyar and Tenenbaum, Joshua B and Schuurmans, Dale and Abbeel, Pieter},
  booktitle=ICLR,
  year={2024}
}

@article{dreamerv3,
  title={Mastering Diverse Domains through World Models},
  author={Hafner, Danijar and Pasukonis, Jurgis and Ba, Jimmy and Lillicrap, Timothy},
  journal={arXiv:2301.04104},
  year={2023}
}

@inproceedings{dreambooth,
  title={{DreamBooth}: Fine Tuning Text-to-Image Diffusion Models for Subject-Driven Generation},
  author={Ruiz, Nataniel and Li, Yuanzhen and Jampani, Varun and Pritch, Yael and Rubinstein, Michael and Aberman, Kfir},
  booktitle=CVPR,
  year={2023}
}

@inproceedings{textualinversion,
  title={An Image is Worth One Word: Personalizing Text-to-Image Generation using Textual Inversion},
  author={Gal, Rinon and Alaluf, Yuval and Atzmon, Yuval and Patashnik, Or and Bermano, Amit H and Chechik, Gal and Cohen-Or, Daniel},
  booktitle=ICLR,
  year={2023}
}

@inproceedings{customdiffusion,
  title={Multi-Concept Customization of Text-to-Image Diffusion},
  author={Kumari, Nupur and Zhang, Bingliang and Zhang, Richard and Shechtman, Eli and Zhu, Jun-Yan},
  booktitle=CVPR,
  year={2023}
}

@inproceedings{adalora,
  title={{AdaLoRA}: Adaptive Budget Allocation for Parameter-Efficient Fine-Tuning},
  author={Zhang, Qingru and Chen, Minshuo and Bukharin, Alexander and He, Pengcheng and Cheng, Yu and Chen, Weizhu and Zhao, Tuo},
  booktitle=ICLR,
  year={2023}
}

@inproceedings{motiondirector,
  title={{MotionDirector}: Motion Customization of Text-to-Video Diffusion Models},
  author={Zhao, Rui and Gu, Yuchao and Wu, Jay Zhangjie and others},
  booktitle=ECCV,
  year={2024}
}

@inproceedings{motionctrl,
  title={{MotionCtrl}: A Unified and Flexible Motion Controller for Video Generation},
  author={Wang, Zhouxia and Yuan, Ziyang and Wang, Xintao and others},
  booktitle={ACM SIGGRAPH},
  year={2024}
}

@inproceedings{clip_radford,
  title={Learning Transferable Visual Models from Natural Language Supervision},
  author={Radford, Alec and Kim, Jong Wook and Hallacy, Chris and Ramesh, Aditya and Goh, Gabriel and Agarwal, Sandhini and Sastry, Girish and Askell, Amanda and Mishkin, Pamela and Clark, Jack and Krueger, Gretchen and Sutskever, Ilya},
  booktitle=ICML,
  year={2021}
}

@inproceedings{dino,
  title={Emerging Properties in Self-Supervised Vision Transformers},
  author={Caron, Mathilde and Touvron, Hugo and Misra, Ishan and J{\'e}gou, Herv{\'e} and Mairal, Julien and Bojanowski, Piotr and Joulin, Armand},
  booktitle=ICCV,
  year={2021}
}

@article{vjepa,
  title={Revisiting Feature Prediction for Learning Visual Representations from Video},
  author={Bardes, Adrien and Garrido, Quentin and Ponce, Jean and Chen, Xinlei and Rabbat, Michael and LeCun, Yann and Assran, Mahmoud and Ballas, Nicolas},
  journal={TMLR},
  year={2024}
}

@inproceedings{r3d18,
  title={A Closer Look at Spatiotemporal Convolutions for Action Recognition},
  author={Tran, Du and Wang, Heng and Torresani, Lorenzo and Ray, Jamie and LeCun, Yann and Paluri, Manohar},
  booktitle=CVPR,
  year={2018}
}

@inproceedings{hoi4d,
  title={{HOI4D}: A 4D Egocentric Dataset for Category-Level Human-Object Interaction},
  author={Liu, Yunze and Liu, Yun and Jiang, Che and others},
  booktitle=CVPR,
  year={2022}
}

@inproceedings{egoschema,
  title={{EgoSchema}: A Diagnostic Benchmark for Very Long-form Video Language Understanding},
  author={Mangalam, Karttikeya and Akshulakov, Raiymbek and Malik, Jitendra},
  booktitle=NIPS,
  year={2023}
}

@article{demsar2006,
  title={Statistical Comparisons of Classifiers over Multiple Data Sets},
  author={Dem{\v s}ar, Janez},
  journal={JMLR},
  year={2006}
}

\clearpage
\beginappendix
\section{Reproducibility Details}
\label{sec:appendix_repro}

\subsection{Hardware and Software}
\label{sec:appendix_hw}

All training and evaluation experiments run on four NVIDIA RTX PRO 6000 Blackwell ($96$\,GB HBM) GPUs in a single host. The software stack is Python 3.12 with CUDA 12.8 and PyTorch 2.11 (cu128 build, including the sm\_120 device kernels required by the Blackwell architecture), together with Diffusers 0.39, PEFT 0.19, and Accelerate 1.13. The two modern metrics that require dedicated environments are isolated. DreamSim is run in a virtual environment with \texttt{transformers}$<$4.50 (required by the official DreamSim distribution). JEDi is run in a separate environment containing the official \texttt{videojedi} package together with the V-JEPA ViT-H/16 feature extractor checkpoint released by the JEDi authors; this environment pins the V-JEPA preprocessing utilities to the versions used in the original ``Beyond FVD'' paper.

\subsection{Hyperparameters}
\label{sec:appendix_hp}

\Cref{tab:hp} lists the hyperparameters shared by the three view-stratified configurations and the second-seed control. All four configurations use identical hyperparameters apart from the training data subset and (for the second-seed control) the random seed.

\begin{table}[h]
\centering
\small
\caption{Adapter training hyperparameters. All four configurations share these values; only the training data subset and (for the seed-137 control) the random seed differ.}
\label{tab:hp}
\begin{tabular}{ll}
\toprule
\textbf{Field} & \textbf{Value} \\
\midrule
Base model & nvidia/Cosmos3-Nano (16B params) \\
LoRA rank $r$ & 32 \\
LoRA scaling $\alpha$ & 64 \\
Target modules & cross-attention $+$ MoE MLPs (gen-modality) \\
Trainable parameters & $87.29{\times}10^6$ ($\sim 0.55\%$ of base) \\
Optimizer & AdamW ($\beta_1{=}0.9$, $\beta_2{=}0.999$, wd $0.01$) \\
Peak learning rate & $3{\times}10^{-4}$ \\
LR schedule & cosine, 100 warm-up steps \\
Optimization steps & 3{,}000 \\
Validation cadence & every 100 steps, $n{=}32$ samples \\
Checkpoint cadence & every 250 steps \\
Selected adapter & iter 1500 (lowest val.\ loss) \\
Effective batch size & 8 (B$=$1, grad acc $=$2, 4 GPUs DDP) \\
Precision & BF16 mixed \\
Activation checkpointing & enabled \\
Video decoder & decord \\
Spatial resolution & $480{\times}832$ \\
Frames per clip & 81 \\
Source FPS & 24 \\
Seed & 42 (and 137 for the combined seed control) \\
\bottomrule
\end{tabular}
\end{table}

\subsection{Inference Settings}
\label{sec:appendix_gen}

For every (configuration, test clip) pair we run image-to-video inference with: 189 frames (Cosmos3-Nano's native maximum, $\sim 7.9$ seconds at 24~FPS), spatial resolution $480{\times}832$, the UniPC multistep scheduler with $\sigma_{\text{shift}}{=}10$, 20 inference steps, classifier-free guidance scale $6$, negative prompt ``\textit{low quality, blurry, distorted, glitches, watermark}'', deterministic seed $42$, output FPS 24.

\subsection{Split Construction}
\label{sec:appendix_splits}

We hold out clips from training and partition the held-out pool into a validation set used for adapter selection ($1{,}388$ clips, from which we sample $n{=}32$ paired clips for the rectified-flow validation loss every 100 training steps) and a test set used for final evaluation ($200$ clips, balanced across both egocentric ($n{=}100$) and exocentric ($n{=}100$) views with rejection sampling for zero overlap with the validation set). The training set ($32{,}105$ clips), the validation set ($1{,}388$ clips), and the test set ($200$ clips) are pairwise disjoint by per-clip unique identifier.

\section{Extended Results}
\label{sec:appendix_results}

\subsection{Exact Validation-Loss $p$-Values}
\label{sec:appendix_extra}

All four LoRA configurations reduce validation diffusion loss on every one
of the $n{=}200$ paired evaluation samples. Under the Wilcoxon signed-rank
test the per-row $p$-values are $\sim10^{-105}$ for egocentric-only,
$\sim10^{-104}$ for exocentric-only, $\sim10^{-104}$ for combined seed-42,
and $\sim10^{-105}$ for combined seed-137. These numbers reflect the
nominal $p$ associated with perfect paired ordering on $n{=}200$ samples
and should be interpreted as ``every paired sample improves over the
base''; we report $p\ll 0.001$ in the main text accordingly.

\subsection{Cross-View Egocentric Transfer Check}
\label{sec:appendix_egoview}

The main test set is a stratified $200$-clip pool covering both views. To isolate the cross-view behaviour of the three adapters, we report paired LPIPS on the egocentric-view subset of the main test set ($n{=}100$ ego clips), so that the same generations used for the main metric panel are reused. \Cref{tab:appendix_ego} shows that all three adapted configurations significantly improve egocentric LPIPS on this subset, with the exocentric-only adapter producing the largest paired improvement ($+0.045$ at $100\%$ paired win-rate) -- i.e., the exocentric-only adapter \emph{does} transfer to the egocentric viewpoint when measured on these clips, and slightly outperforms the egocentric-only adapter ($+0.039$ at $96\%$ paired win-rate). The combined configuration is comparable to ego-only ($+0.038$ at $98\%$). This pattern is consistent with the headline result: under matched optimization budget, the exocentric subset is doing the heavy lifting of adaptation, including at cross-view test time.

\begin{table}[h]
\centering
\small
\caption{Cross-view check on the $100$-clip egocentric subset of the main test set. All three adapted configurations significantly improve egocentric LPIPS on this subset; the exocentric-only adapter delivers the largest improvement, supporting the headline finding that exocentric-only training is a strong default in this corpus.}
\label{tab:appendix_ego}
\begin{tabular}{lccc}
\toprule
\textbf{Configuration} & \textbf{LPIPS $\Delta$} & \textbf{Win} & \textbf{$p$ (Wilcoxon)} \\
\midrule
Egocentric-only & $+0.039$ & $96.0\%$ & $<10^{-13}$ \\
Exocentric-only & $\mathbf{+0.045}$ & $\mathbf{100\%}$ & ${<}\mathbf{10^{-14}}$ \\
Combined        & $+0.038$ & $98.0\%$ & $<10^{-13}$ \\
\bottomrule
\end{tabular}
\end{table}

\subsection{Rollout-Horizon Curve}
\label{sec:appendix_time}

\begin{figure}[h]
\centering
\includegraphics[width=\linewidth]{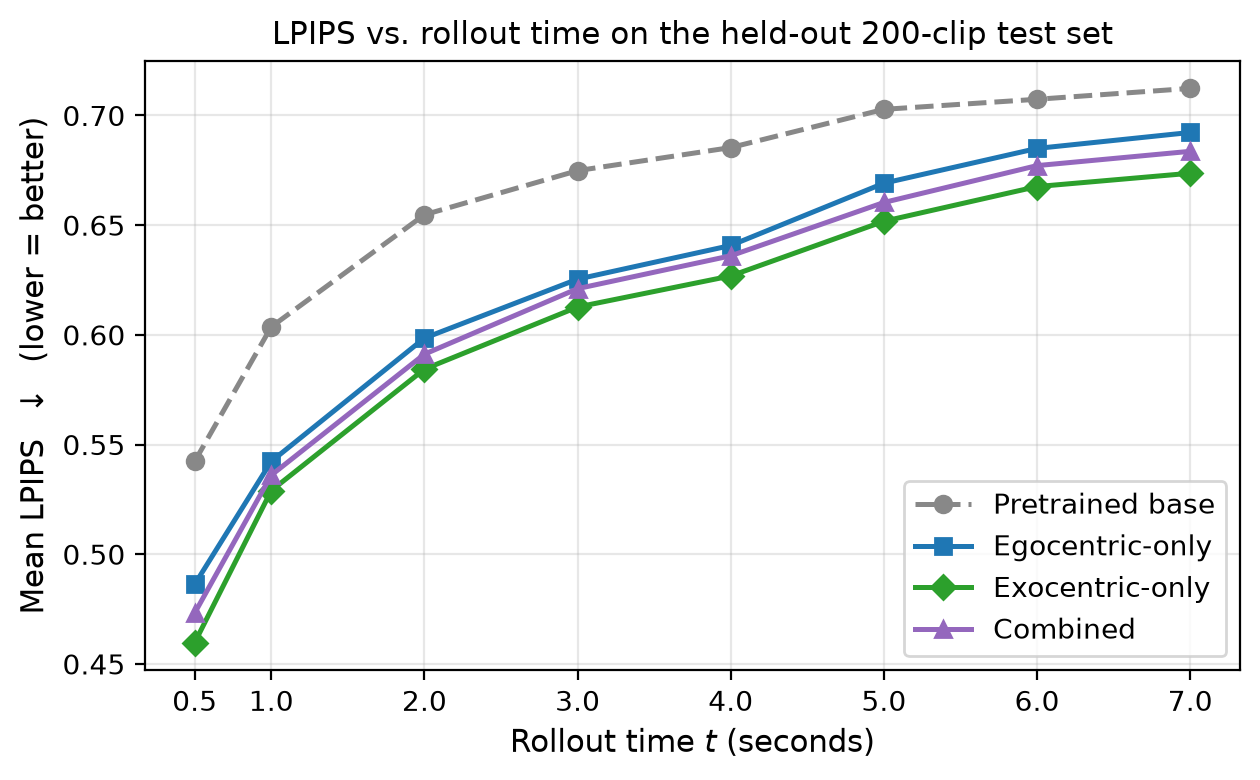}
\caption{Mean LPIPS as a function of rollout time on the held-out 200-clip test set, one curve per configuration. The exocentric-only configuration retains the lowest LPIPS at every time stamp. The absolute gap to the pretrained base is widest at the smallest rollout times and narrows as $t$ grows, since all configurations drift away from the single ground-truth trajectory.}
\label{fig:lpips_vs_time}
\end{figure}

\Cref{fig:lpips_vs_time} plots mean LPIPS as a function of rollout time on the held-out 200-clip test set, one curve per configuration. All four curves rise monotonically with $t$: at $t{=}0.5$\,s the base mean LPIPS is $0.542$ versus $0.460$ for the exocentric-only configuration (a gap of $0.082$); at $t{=}4.0$\,s the base reaches $0.685$ versus exo $0.627$ (a smaller gap of $0.058$). The absolute gap to the pretrained base is widest at the smallest rollout times and narrows steadily as $t$ grows, since all configurations drift away from the single ground-truth trajectory in the same way. The exocentric-only configuration retains the smallest LPIPS at every time stamp.

\section{Practical Notes}
\label{sec:appendix_notes}

\subsection{R3D-Fréchet vs.\ Canonical I3D-FVD}
The canonical Fréchet video distance is computed on features extracted by the Inflated 3D ConvNet (I3D) of \cite{i3d}. A turnkey I3D feature extractor with Kinetics-400 weights is not currently packaged with mainstream deep-learning frameworks; for this paper we use the closely related ResNet-3D-18 video classifier (\texttt{torchvision.models.video.r3d\_18} with \texttt{KINETICS400\_V1} weights) and refer to the resulting metric as R3D-Fréchet to make the deviation explicit. Both feature extractors are pretrained for Kinetics-400 action classification at the same input clip length, so the Fréchet distance retains its intended interpretation as a measure of distributional similarity in a Kinetics-aware embedding. Absolute values, however, are not directly comparable to I3D-FVD numbers from the broader literature. Recomputing the Fréchet-style metric with the canonical I3D backbone is straightforward but requires installing the I3D feature extractor outside the standard framework distribution; we recommend doing so for any direct comparison with prior work.

\subsection{Time-Aligned Sampling}
\label{sec:appendix_time_align}
A frame-index-based implementation of LPIPS would mis-align Cosmos3-Nano's 24~FPS generated output against the 4~FPS ground-truth clips, yielding pairings of frames separated by up to two seconds of wall-clock time. Wall-clock time-aligned sampling (16 evenly-spaced timestamps in $[0,\min(\text{gen\_dur},\text{gt\_dur})]$, dropping $t{=}0$ to remove the trivial first-frame match) increases the paired LPIPS gap on our test set by approximately $3.9\times$ over the index-based comparison. We recommend the time-aligned protocol for any video evaluation where the generated and reference frame rates differ.

\subsection{Adapter-Merge Verification}
\label{sec:appendix_merge}
For inference we explicitly merge the trained adapter $\Delta W = (\alpha/r)\,BA$ into each corresponding base weight matrix and verify the merge numerically. Across all $36$ transformer blocks the merged weight differs from the base in $98.83\%$ of its entries, and the per-tensor maximum magnitude of $\Delta W$ reaches $17.8\%$ of the base weight magnitude. This sanity check rules out a class of silent-failure modes (adapter loading as a zero increment) that would otherwise produce numerically valid but functionally untrained outputs.

\section{Release Policy and Per-Clip Data Availability}
\label{sec:appendix_data}

\nbf{Release} The release accompanying this paper will include a representative subset of RetailSMV sufficient to reproduce the reported metrics, the full set of evaluation scripts (data preparation, generation, metric computation, paired statistical tests), the generated videos used in the main results, and all per-clip metric files. Full corpus access is controlled: requests will be served on a case-by-case basis through the same channel as the released subset, subject to the privacy and licensing terms under which the corpus was collected (\cref{sec:dataset}).

\nbf{Per-clip data} For each (configuration, test clip) pair we provide per-clip time-aligned LPIPS, PSNR, SSIM, Hessel CLIPScore, and DreamSim, together with the per-clip wall-clock LPIPS series at $t\in\{0.5,1.0,2.0,3.0,4.0,5.0,6.0,7.0\}$ seconds. We also provide the $800$ generated videos (four configurations $\times$ $200$ test clips), the precomputed R3D feature arrays used to compute R3D-Fréchet, and the V-JEPA ViT-H/16 feature arrays used to compute JEDi (one $(B{,}D)$ array per configuration plus the ground-truth feature array).

\section{Contributions and Acknowledgments}
\label{sec:appendix_contributors}

\nbf{Core Contributors} Amirreza Rouhi, Rajat Aggarwal, Parikshit Sakurikar, Anoop M.\ Namboodiri, Sashi P.\ Reddi.

\nbf{Contact} Correspondence and corpus-access requests can be directed to the core contributors at the following addresses:
\begin{itemize}[leftmargin=*,nosep]
  \item \texttt{amir@dreamvu.ai}
  \item \texttt{rajat@dreamvu.ai}
  \item \texttt{parikshit@dreamvu.ai}
  \item \texttt{anoop@dreamvu.ai}
  \item \texttt{sashi@dreamvu.ai}
\end{itemize}

\end{document}